\documentclass[letterpaper,10pt,conference]{ieeeconf}
\IEEEoverridecommandlockouts       
\usepackage{amssymb}
\usepackage{amsmath}
\usepackage[english]{babel}
\usepackage{float}
\usepackage{graphicx}
\usepackage{hyperref}
\usepackage{mathtools}

\usepackage{filecontents}
\usepackage[noadjust]{cite}
\usepackage[font=footnotesize,labelfont=footnotesize]{subcaption}
\usepackage[font=footnotesize,labelfont=footnotesize]{caption}
\usepackage{comment}
\usepackage{authblk}
\usepackage{multirow}
\usepackage{xcolor}
\usepackage{algorithm}
\usepackage{algorithmic}
\usepackage{xspace}

\DeclareMathOperator*{\argmin}{arg\,min}
\DeclareMathOperator*{\argmax}{arg\,max}

\definecolor{darkblue}{rgb}{0.15,0.15,0.55}
\definecolor{lightgrey}{rgb}{0.75,0.75,0.75}

\providecommand{\codecomment}[1]{\textcolor{lightgrey}{\dotfill}\textcolor{darkblue}{//\,\textrm{#1}}}
\providecommand{\codecommentline}[1]{\textcolor{darkblue}{\textbackslash*\,\textrm{#1}*\textbackslash\,}}

\newcommand{\nphard}{$\mathcal{NP}$-hard\xspace}

\newcommand{\thm}{\noindent \textbf{Theorem}\xspace}
\newcommand{\lem}{\noindent \textbf{Lemma}\xspace}

\newcommand{\pf}{\noindent \textbf{Proof}\xspace}

\newcommand{\qed}{\hfill $\square$}

\begin{document}
\title{\LARGE \bf Planning for target retrieval using a robotic manipulator in cluttered and occluded environments}
\author{Changjoo Nam, Jinhwi Lee, Younggil Cho, Jeongho Lee, Dong Hwan Kim, and ChangHwan Kim$^*$
\thanks{This work was supported by the Technology Innovation Program and Industrial Strategic Technology Development Program (10077538, Development of manipulation technologies in social contexts for human-care service robots). The authors are with Korea Institute of Science and Technology.
E-mail: {\tt\small\{cjnam, jinhooi, briancho, kape67, gregorykim, ckim\}@kist.re.kr}. $^*$Corresponding author}\\
}

\maketitle

\begin{abstract}
This paper presents planning algorithms for a robotic manipulator with a fixed base in order to grasp a target object in cluttered environments. We consider a configuration of objects in a confined space with a high density so no collision-free path to the target exists. The robot must relocate some objects to retrieve the target while avoiding collisions. For fast completion of the retrieval task, the robot needs to compute a plan optimizing an appropriate objective value directly related to the execution time of the relocation plan.

We propose planning algorithms that aim to minimize the number of objects to be relocated. Our objective value is appropriate for the object retrieval task because grasping and releasing objects often dominate the total running time. In addition to the algorithm working in fully known and static environments, we propose algorithms that can deal with uncertain and dynamic situations incurred by occluded views. The proposed algorithms are shown to be complete and run in polynomial time. Our methods reduce the total running time significantly compared to a baseline method (e.g., 25.1\% of reduction in a known static environment with 10 objects). 

\end{abstract}


\section{Introduction}
\begin{figure}[t]
    \captionsetup{skip=0pt}
    \centering
	\includegraphics[width=0.48\textwidth]{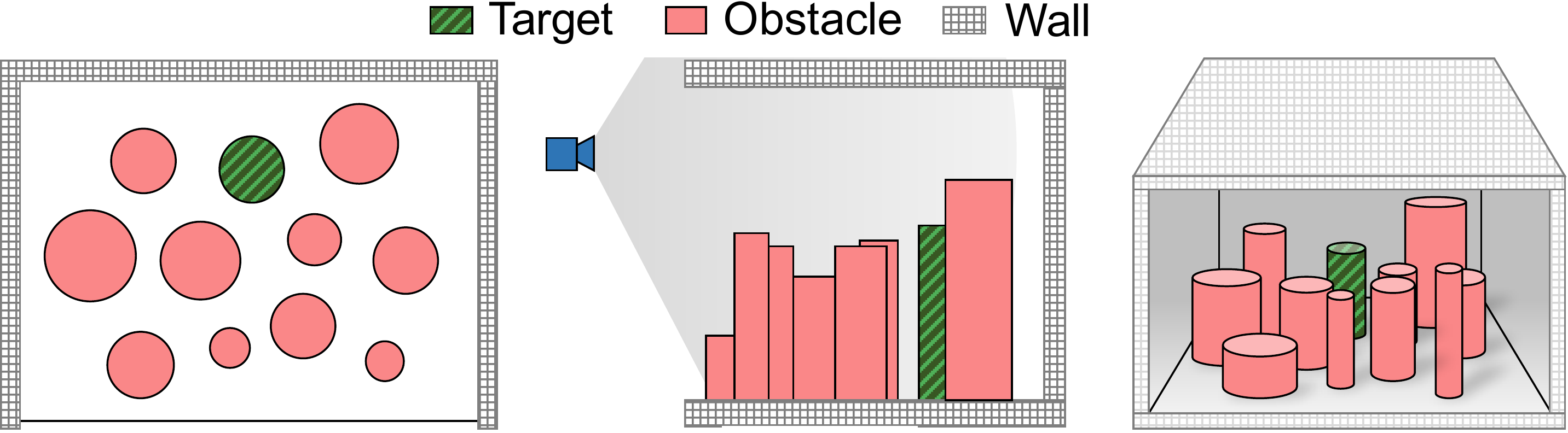}
    \caption{Objects in dense clutter. The target (green stripes) and some obstacles (pink solid) are occluded. An object could be fully or partially occluded and inaccessible to the robot without relocating some front objects.}
    \label{fig:example}
    \vspace{-10pt}
\end{figure}

\vspace{-2pt}
Retrieving a target object from clutter using a robotic manipulator has long been considered as an important and practical task. Robots will perform such tasks in cluttered and confined spaces frequently in our home or workplace (e.g., shelves in fridges and cupboards, item pods in Amazon's warehouse) as illustrated in Fig.~\ref{fig:example}. If obstacles surrounding a target object are populated densely, it is necessary to rearrange a subset of the obstacles to achieve the mission retrieving the target without collisions. However, planning for obstacle rearrangement has shown to be \nphard even in fully known and static environments~\cite{wilfong1991motion,stilman2008planning}.
Cluttered and confined environments often produce uncertainties owing to the occlusions incurred by objects at the front. The robot may have to plan without locating some objects including the target.

\begin{figure}[t]
    \captionsetup{skip=0pt}
    \centering
	\includegraphics[width=0.33\textwidth]{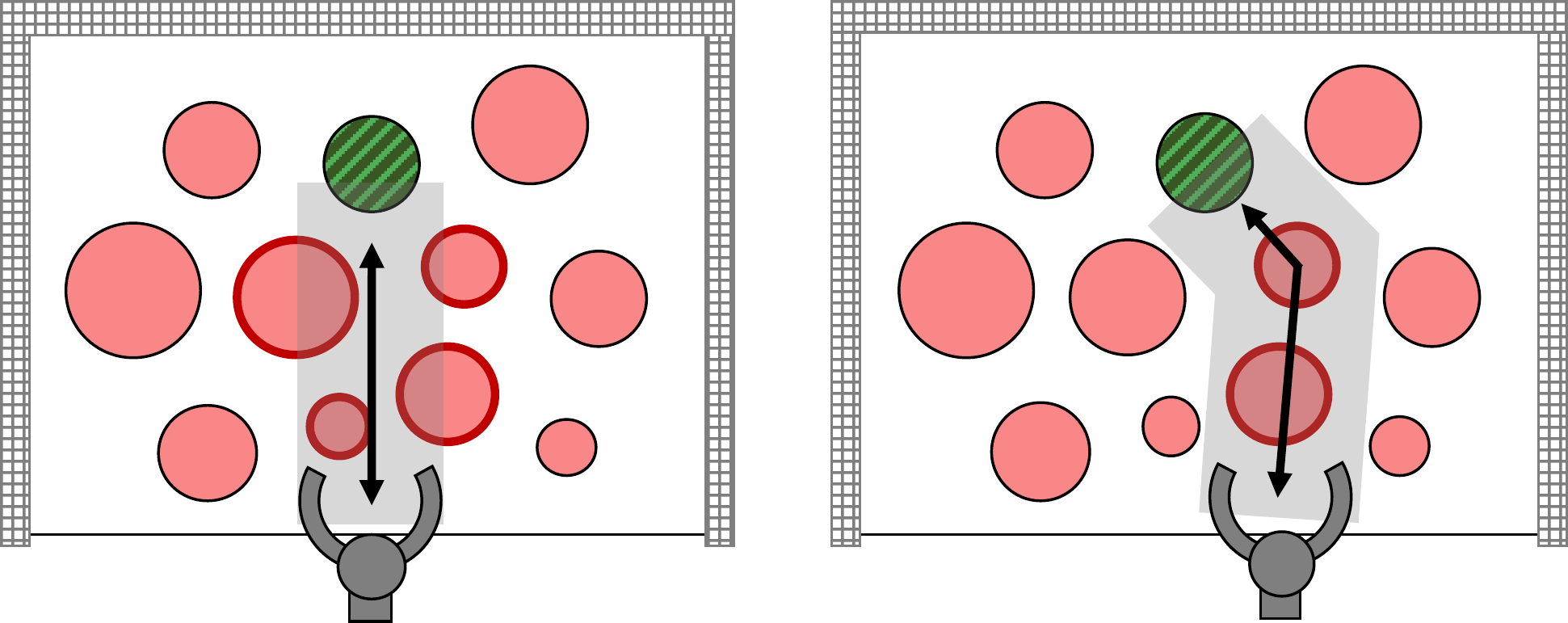}
    \caption{An example demonstrating the importance of choosing an appropriate objective value for relocation planning. (L) The shortest path to the target needs relocation of four obstacles (red bold outlines). (R) The path minimizing the number of relocated obstacles needs relocation of two.}
  \label{fig:comparison}
\vspace{-20pt}
\end{figure}

Although there have been numerous methods presented to manipulate a target object in clutter~\cite{srivastava2014combined,dogar2012planning,havur2014hybrid,garrett2015backward,haustein2015kinodynamic,moll2018randomized}, the focus has been on finding valid grasps/paths or incorporating non-prehensile actions, but not on the optimization of relocation plans. Fig.~\ref{fig:comparison} shows an illustrative example showing the importance of optimizing relocation plans with an appropriate objective value. The target (green stripes) is surrounded by obstacles (pink solid) and walls. Since the space is confined, the manipulator only can navigate through the bottom side of the space. If the manipulator takes the shortest path to the target (Fig.~\ref{fig:comparison}-L) as done in~\cite{dogar2012planning}, the four obstacles on the path (red bold outlines) should be removed. On the other hand, the number of relocated obstacles reduces to two if the manipulator aims to minimize the number of relocation (Fig.~\ref{fig:comparison}-R). This longer path takes a shorter execution time than the distance-optimal path since grasping and releasing objects dominate the total execution time while just transporting objects takes relatively little time.

In this work, we propose planning algorithms for robotic manipulators with a fixed base in order to retrieve a target object from clutter. We assume that objects are densely located thus no collision-free path of the end-effector to the target exists without relocating some objects. The algorithms compute a plan while minimizing the number of objects to be relocated. We begin from considering a fully known environment. Then, we take into account uncertain and dynamic situations arising from occlusions. 

The following are contribution of this work:\vspace{-2pt}
\begin{itemize}
    \item We propose a polynomial-time and complete algorithm that computes a relocation plan in a known static environment. We extend the algorithm to consider uncertain and dynamic situations caused by occlusion (Sec.~\ref{sec:alg}).
    \item We provide mathematical proofs for the time complexity and completeness of the proposed algorithms (Sec.~\ref{sec:alg}).
    \item We show results from extensive simulations and experiments in various scenarios with a comparison. We also run experiments using a physical robot integrated with a vision system (Sec.~\ref{sec:exp}).
\end{itemize}

\section{Related Work}
\label{sec:related}
\vspace{-2pt}
The work presented in~\cite{dogar2012planning} proposes a planning framework to grasp a target in cluttered and known environments. It removes obstacles that are in the shortest path of the end-effector to the target (like Fig.~\ref{fig:comparison}-L). Although this method finds the distance-optimal path, some obstacles could have to be removed unnecessarily since the objective value is not the number of obstacles to be removed. Other works, such as \cite{srivastava2014combined,haustein2015kinodynamic,moll2018randomized}, also do not directly optimize the relocation plan but mainly concern about validity of the plan.

Some recent work considers partially known environments. The algorithm proposed in~\cite{dogar2014object} computes a sequence of objects to be removed while minimizing the expected time to find a hidden target. The strength of this work is the mathematical formalization of the search and grasp planning problem. However, the algorithm shows exponential running time so may not be practically useful in environments with densely packed objects. In the experiment with five objects, planning takes longer than 25\,sec.
Another work~\cite{lin2015planning} finds a sequence of actions of a mobile manipulator that minimizes the expected time to reveal all possible hidden target poses. This work defines admissible costs for its A$^*$ search, but planning takes long time owing to the high branching factor of the search (e.g., 40\,sec with five objects). There have been several approaches~\cite{nie2016searching,li2016act} modeling the problem as a Partially Observable Markov Decision Process but they do not seem to scale even with moderate-sized instances.
 

Among these, no work has formulated the problem as an optimization problem whose objective value is the number of obstacles to be relocated. The methods presented in these work require substantial planning time in clutter. The examples that we will consider are significantly more cluttered so we need faster planning algorithms. In our own work~\cite{lee2019efficient_arxiv}, we present a fast algorithm for relocation in known environments by employing a collision avoidance method called Vector Field Histogram+ (VFH+)~\cite{ulrich1998vfh+}. Although it shows good performance in dense clutter, it does not aim to find a global optimal solution since VFH+ is a local planning method which focuses on the vicinity of the target but not the entire space. The present work sets out to achieve the global optimum and considers partially known environments.

\section{Problem Formulation}
\label{sec:def}
\vspace{-2pt}
Target retrieval from clutter requires several different processes such as perception, relocation planning, grasping, path planning, and navigation. We focus on the relocation planning in order to generate a collision-free path of the end-effector of the robot. 
The problem of finding a path in a configuration of movable objects has an exponentially large search space in the number of movable objects. A simplified version of the problem with only one movable object is shown to be \nphard~\cite{wilfong1991motion,stilman2008planning} even in a perfectly known environment. We begin from a known environment and extend the scope to include uncertainties incurred by occlusions. In this section, we describe the assumptions, the problem definition, and the environments.

\subsection{Assumptions}
\vspace{-2pt}
We assume that all objects are densely populated in a workspace of a robotic manipulator. No collision-free path exists for the end-effector without relocating some objects. 
For simplicity, we model objects as 3D cylindrical structures, so the objects can be grasped from any direction. Irregularly shaped objects which could have restricted reachable directions for grasping will be considered in our future work. We consider grasping objects on the side along a 2D path (with a fixed height in the 3D space) but do not consider picking objects from the top which enables grasping the target without relocation. 
We consider a fixed top-front camera view whose field of view is wide enough to capture the entire workspace of the robot. 
Using a fixed view camera does not make the problem easier since mobile cameras could acquire more information from various view points.

\subsection{Problem definition}
\begin{figure}[h!]
  \begin{center}
    \begin{minipage}{0.4\linewidth}
      \includegraphics[scale=0.4]{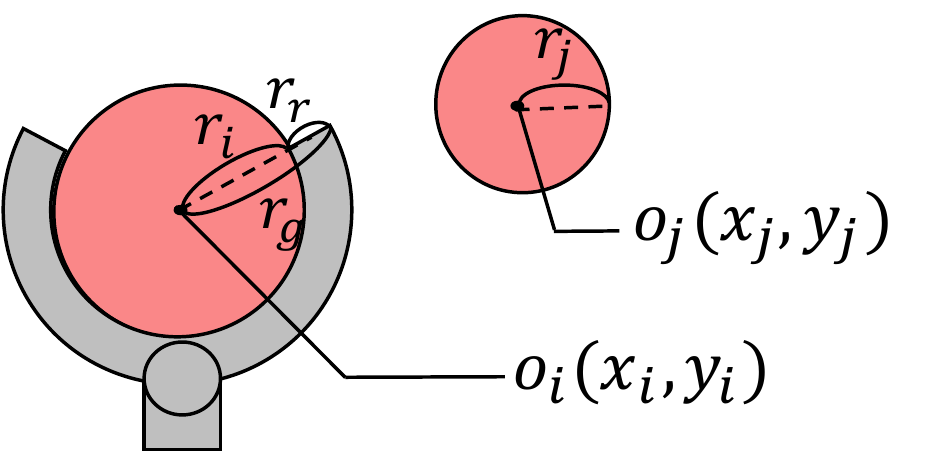}
    \end{minipage}\quad
    \begin{minipage}{0.4\linewidth}
	\caption{Object $i$ at $(x_i, y_i)$ with a radius $r_i$. If $o_i$ grasped, the size of the end-effector $r_r$ is added to $r_i$.}\label{fig:geo}
    \end{minipage}
    \vspace{-10pt}
  \end{center}
  \vspace{-5pt}
\end{figure}

\vspace{-2pt}
We set out to minimize the total running time to complete the mission, which includes planning time and execution (manipulation) time. Since grasping and releasing dominate the execution time, we pose the problem as a minimization problem where the objective value is the number of objects to be relocated. Suppose that a configuration is with obstacles $o_i$ for $i = 1, \cdots, N \in \mathbb{Z}^{+}$ and a target $o_t$ (so total $N+1$ objects). The centroid, radius, and height of object $i$ is described by $(x_i, y_i)$, $r_i$, and $h_i$, respectively. The set $O$ includes all objects so $O = \{o_1, \cdots, o_N, o_t\}$. Let $O_R \subset O$ be the sequence of obstacles to be relocated (excluding the target) where $|O_R| = k \le N$. 
The base of the manipulator and the camera are fixed at $(x_r, y_r, h_r)$ and $(x_c, y_c, h_c)$, respectively. The thickness of the end-effector is $r_r$. If it grasps an object whose radius is $r_i$, the radius of the end-effector grasping the object is $r_g = r_i + r_r$. Fig.~\ref{fig:geo} shows the geometry of objects and the end-effector.\vspace{-8pt}

A mathematical definition of the problem is to find $O_R$ that minimizes $k$. The solution sequence $O_R$ lists obstacles in the order in which they should be removed.

\subsection{Dynamic and uncertain situations owing to occlusion}
\label{sec:cases}
\begin{figure}[h!]
\vspace{-5pt}
    \captionsetup{skip=0pt}
    \centering
    \begin{subfigure}{0.134\textwidth}
    \captionsetup{skip=0pt}
	    \includegraphics[width=\textwidth]{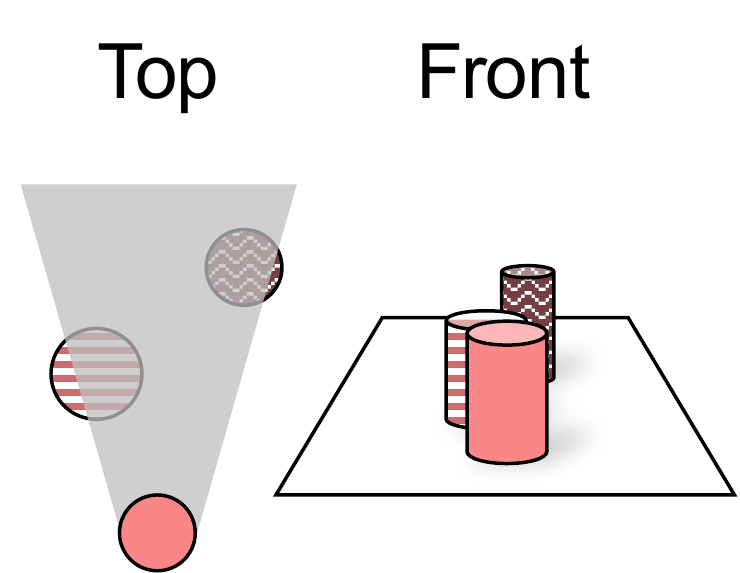}
        \caption{Occluded objects}
        \label{fig:o_obj}
    \end{subfigure}%
    \begin{subfigure}{0.21\textwidth}
    \captionsetup{skip=0pt}
	    \includegraphics[width=\textwidth]{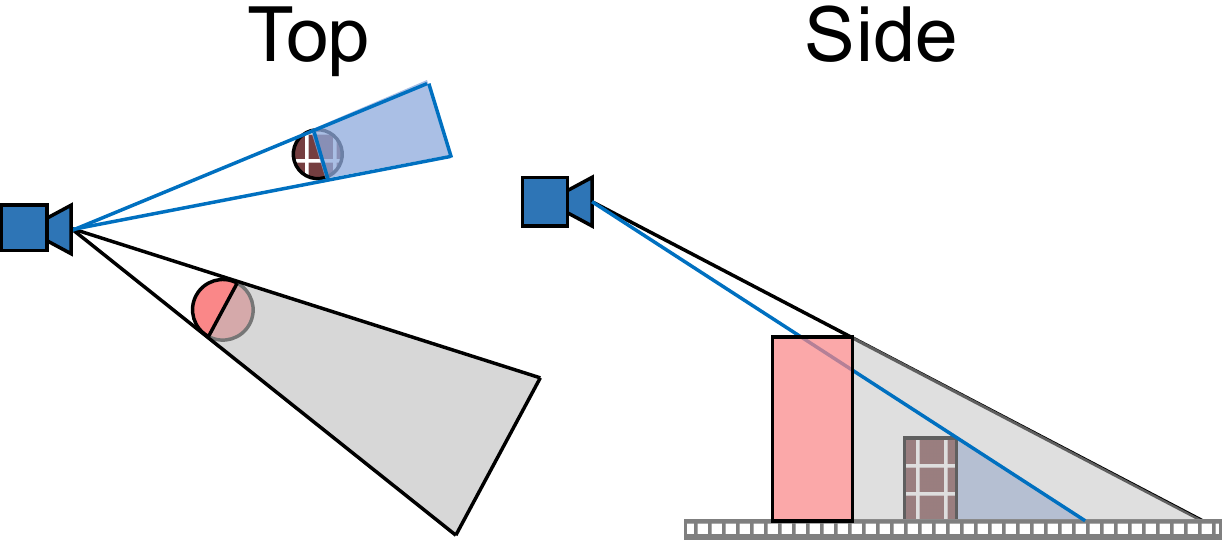}
        \caption{Occluded volume (shades)}
        \label{fig:o_vol}
    \end{subfigure}
    \begin{subfigure}{0.13\textwidth}
        \captionsetup{skip=0pt}
	    \includegraphics[width=\textwidth]{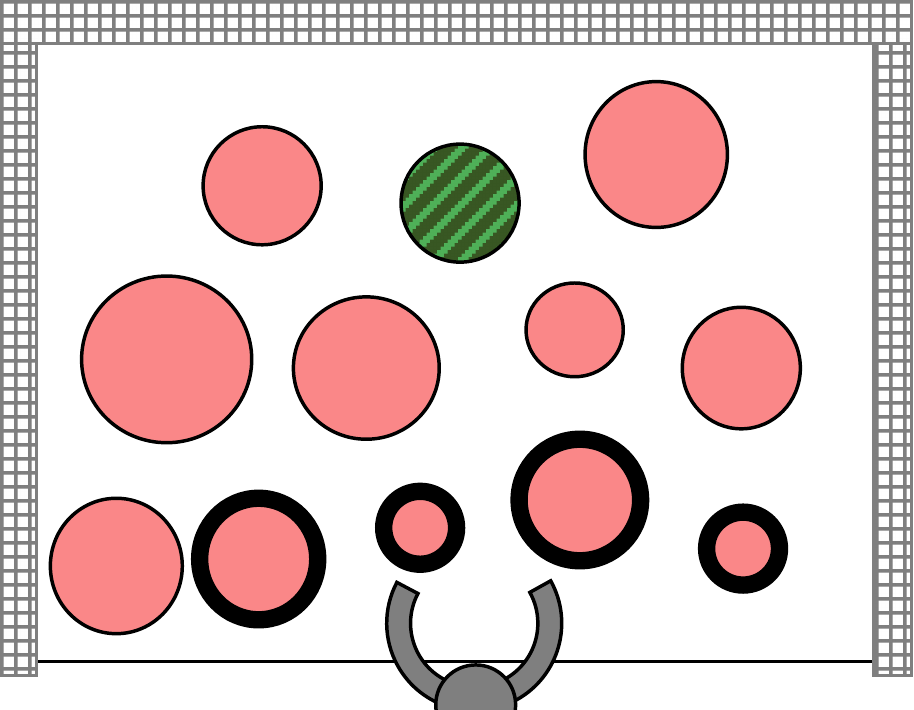}
        \caption{Accessible objects (bold outlines)}
        \label{fig:acc}
  \end{subfigure}
    \caption{Concepts related to occlusion in clutter.}
  \label{fig:defs}
\vspace{-18pt}
\end{figure}

\vspace{-2pt}
Objects could occlude each other in dense clutter. We need to consider different situations occurring from occlusions so define relevant concepts. An object is \textit{occluded} if it is partially visible to the robot. \textit{Occluded volume} quantifies the space occluded by objects (Fig.~\ref{fig:o_vol}). An object is \textit{accessible} if it can be grasped by the end-effector without relocating any objects. The set $O_A \subset O$ includes all accessible objects. We assume that $O_A \neq \emptyset$. Fig.~\ref{fig:acc} shows four accessible objects (bold outlines). 

\begin{figure}[h!]
\vspace{-10pt}
    \captionsetup{skip=0pt}
    \centering
	\includegraphics[width=0.4\textwidth]{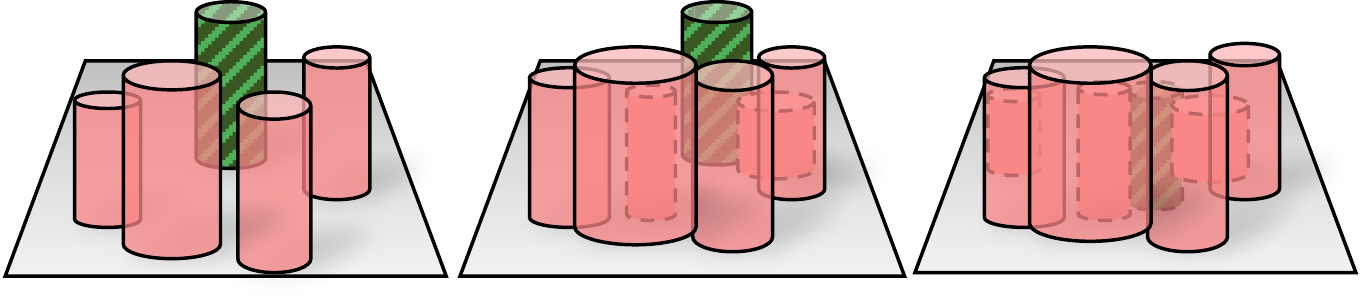}
    \caption{(L) Base case: All objects including the target (green stripes) are known. (C) Case I: Some objects excluding the target are unknown (dotted outlines). (R) Case II: Some objects including the target are unknown.}
  \label{fig:uncertain}
\vspace{-10pt}
\end{figure}

We consider three cases illustrated in Fig.~\ref{fig:uncertain}. (Base case) Known geometry of $O$ and detected target: $O_R$ can be computed before the robot starts relocation. (Case I) Partially known geometry of $O$ and detected target: some completely occluded obstacles appear dynamically while the robot executes an initially computed plan. (Case II) Partially known geometry of $O$ and undetected target: some completely occluded objects including the target appear dynamically while the robot relocates objects. 

\section{Planning algorithms for object relocation}
\label{sec:alg}
\vspace{-2pt}
We describe task planning algorithms for the three cases described in Sec.~\ref{sec:cases} and provide analyses of them. 

\subsection{Base case: Planning with full information}
\label{sec:static}
\vspace{-2pt}
We develop an algorithm for Base case that (i) constructs a graph representing the configuration of objects and then (ii) finds the minimum-hop path on the graph, which represents the sequence of obstacles to be relocated. 

\begin{figure}[h!]
\vspace{-5pt}
    \captionsetup{skip=0pt}
    \centering
   \begin{subfigure}{0.097\textwidth}
   \captionsetup{skip=0pt}
	\includegraphics[width=\textwidth]{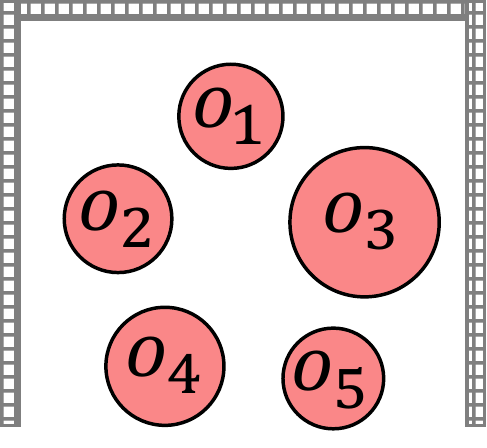}
	\caption{}
    \label{fig:idea1}
  \end{subfigure}%
  \begin{subfigure}{0.097\textwidth}
  \captionsetup{skip=0pt}
	\includegraphics[width=\textwidth]{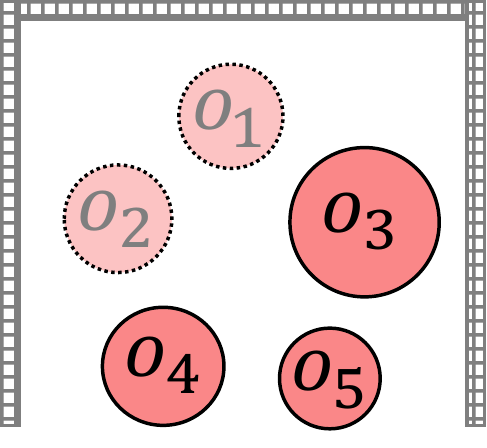}
	\caption{}
    \label{fig:idea2}
  \end{subfigure}%
  \begin{subfigure}{0.097\textwidth}
  \captionsetup{skip=0pt}
	\includegraphics[width=\textwidth]{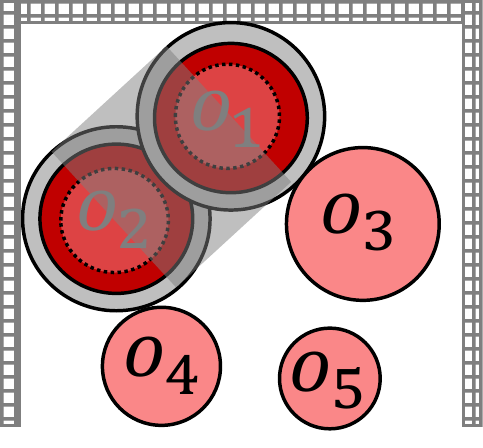}
	\caption{}
    \label{fig:idea3}
  \end{subfigure}%
 \begin{subfigure}{0.097\textwidth}
   \captionsetup{skip=0pt}
	\includegraphics[width=\textwidth]{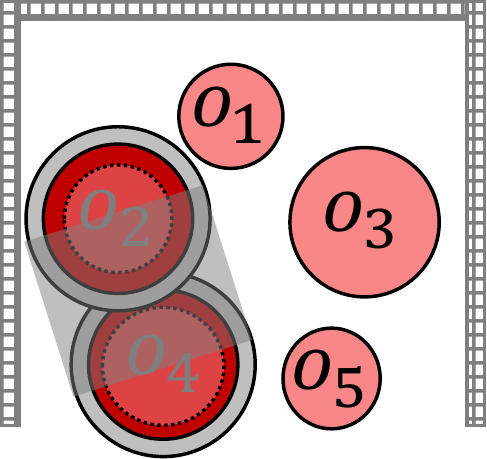}
	\caption{}
    \label{fig:idea4}
  \end{subfigure}%
 \begin{subfigure}{0.097\textwidth}
   \captionsetup{skip=0pt}
	\includegraphics[width=\textwidth]{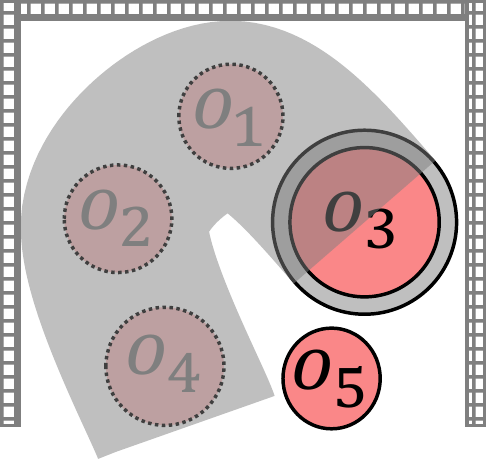}
	\caption{}
    \label{fig:idea5}
  \end{subfigure}
  \caption{A path for target retrieval. (a) An initial configuration. (b) Suppose that $o_1$ and $o_2$ do not exist. (c) If the end-effector grasping the largest object $o_3$ (the gray ring adds the end-effector size) can move between the poses of $o_1$ and $o_2$ without collision, a path exists between the two poses. (d) The same applies to the path between $o_2$ and $o_4$. (e) An example trajectory that $o_3$ can be retrieved from the clutter if the objects on the path are removed.}
  \label{fig:idea}
\vspace{-10pt}
\end{figure}

\smallskip
\noindent \textit{Graph construction: }
We construct a graph representing movable paths of objects. The basic idea is illustrated in Fig.~\ref{fig:idea}. An edge between a pair of nodes means a collision-free path for the end-effector to move any object between the two poses represented by the nodes. Fig.~\ref{fig:idea1} shows a configuration of the five objects. Suppose that $o_1$ and $o_2$ do not exist (Fig.~\ref{fig:idea2}). In Fig.~\ref{fig:idea3}, the largest object $o_3$ grasped by the robot is dilated by the end-effector size $r_r$ (gray ring). If the end-effector grasping the largest object can move between the poses of $o_1$ and $o_2$ without collisions (the gray shade represents the trajectory of the end-effector), it can move any object through the same path since $o_3$ is the largest. Then an edge is connected between the two nodes representing $o_1$ and $o_2$. The same applies to the path between $o_2$ and $o_4$ (Fig.~\ref{fig:idea4}). If $o_4$, $o_2$, and $o_1$ are removed sequentially, $o_3$ can be retrieved (Fig.~\ref{fig:idea5}). Similarly, removing only $o_5$ also enables the robot to retrieve $o_3$.

\begin{figure*}[h!]
    \captionsetup{skip=-1pt}
    \centering
   \begin{subfigure}{0.21\textwidth}
   \captionsetup{skip=-1pt}
	\includegraphics[width=\textwidth]{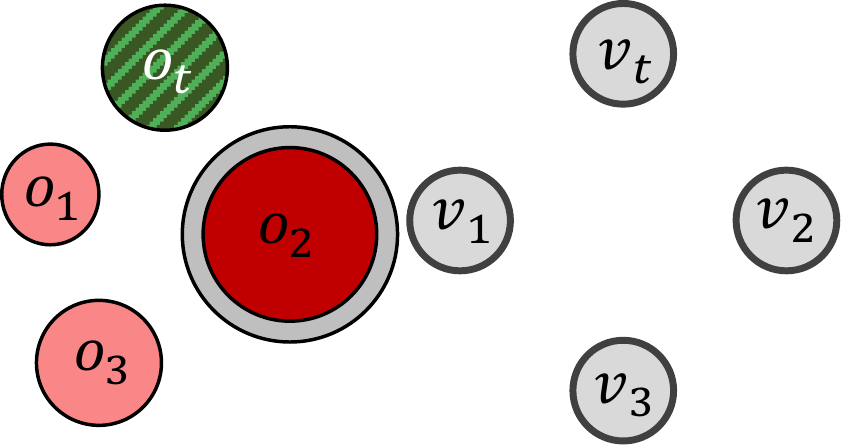}
    \caption{}
    \label{fig:graph1}
  \end{subfigure}
  \begin{subfigure}{0.21\textwidth}
  \captionsetup{skip=-1pt}
	\includegraphics[width=\textwidth]{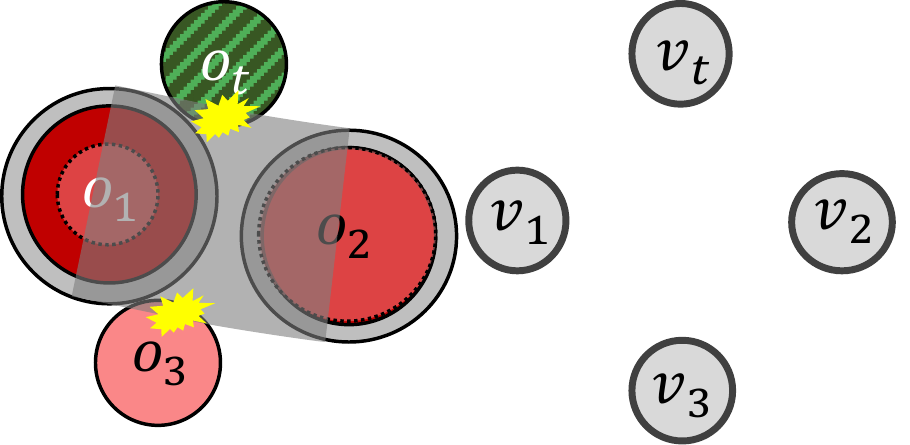}
    \caption{}
    \label{fig:graph2}
  \end{subfigure}
  \begin{subfigure}{0.21\textwidth}
  \captionsetup{skip=-1pt}
	\includegraphics[width=\textwidth]{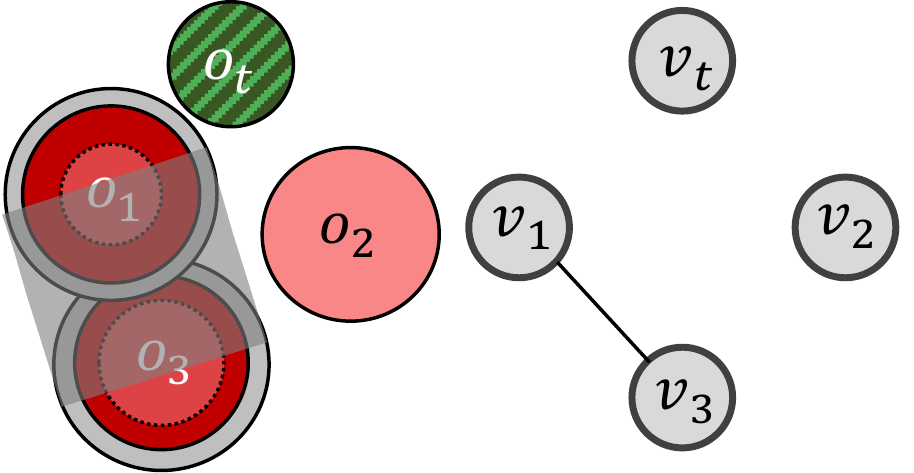}
    \caption{}
    \label{fig:graph3}
  \end{subfigure}
  \begin{subfigure}{0.21\textwidth}
  \captionsetup{skip=-1pt}
	\includegraphics[width=\textwidth]{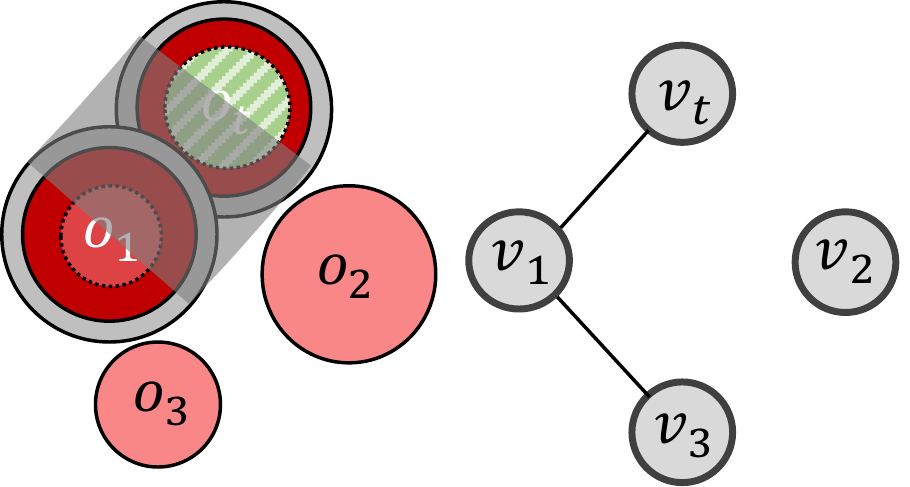}
    \caption{}
    \label{fig:graph4}
  \end{subfigure}\\ \vspace{-5pt}
  \begin{subfigure}{0.21\textwidth}
   \captionsetup{skip=-1pt}
	\includegraphics[width=\textwidth]{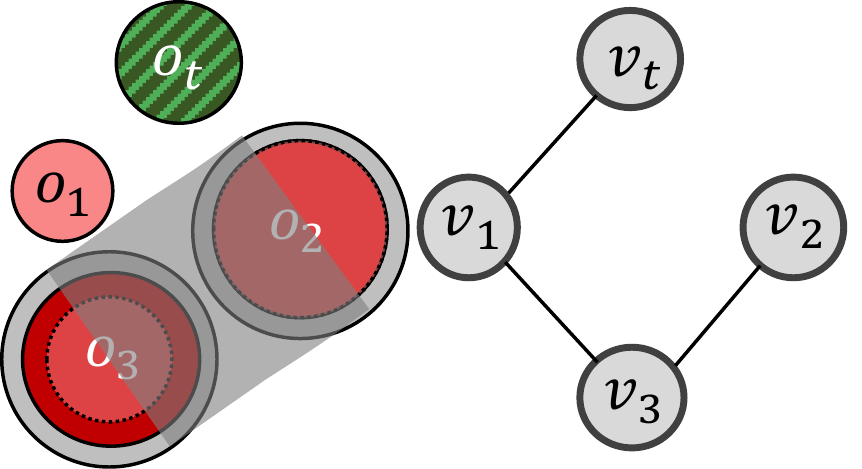}
    \caption{}
    \label{fig:graph5}
  \end{subfigure}
  \begin{subfigure}{0.21\textwidth}
  \captionsetup{skip=-1pt}
	\includegraphics[width=\textwidth]{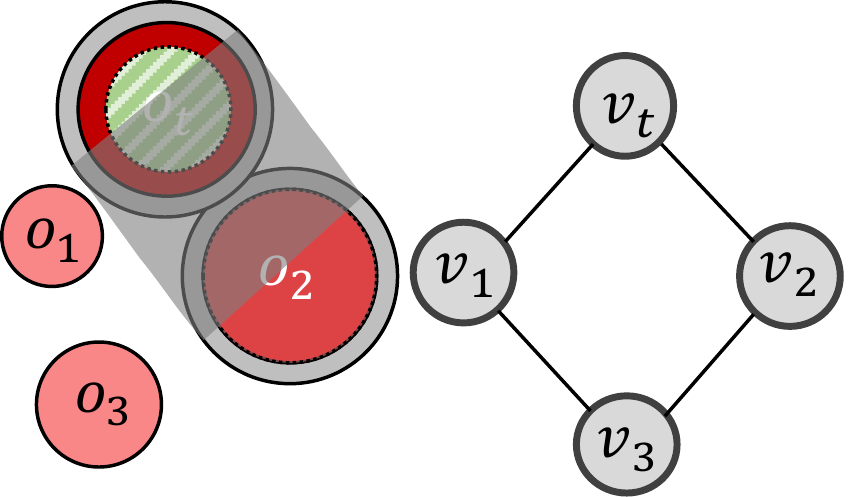}
    \caption{}
    \label{fig:graph6}
  \end{subfigure}
  \begin{subfigure}{0.21\textwidth}
  \captionsetup{skip=-1pt}
	\includegraphics[width=\textwidth]{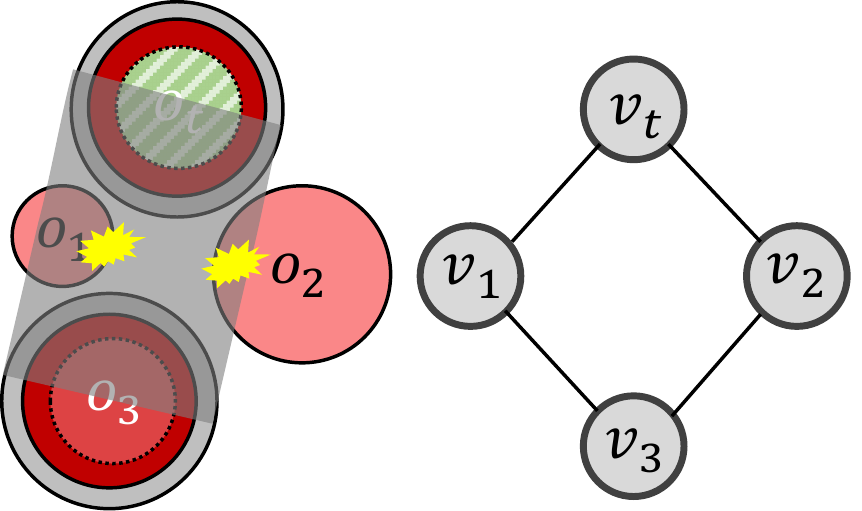}
    \caption{}
    \label{fig:graph7}
  \end{subfigure}
  \begin{subfigure}{0.21\textwidth}
  \captionsetup{skip=-1pt}
	\includegraphics[width=\textwidth]{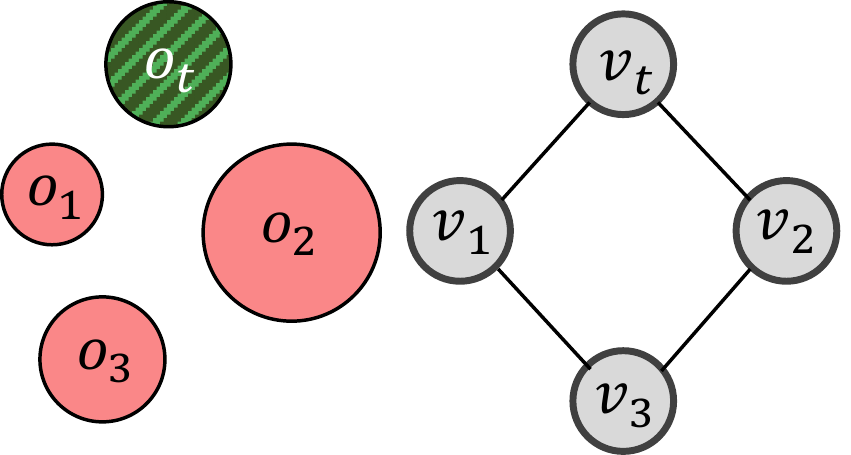}
    \caption{}
    \label{fig:graph_final}
  \end{subfigure}
  \caption{An example of graph construction. Walls around the objects (like Fig.~\ref{fig:idea}) are omitted. (a) Nodes are generated from $O$ and the largest object ($o_2$ in dark red) is dilated by the robot size and a safety margin (gray ring). (b) The largest object cannot move between $(x_1, y_1)$ and $(x_2, y_2)$, so nodes $v_1$ and $v_2$ are not connected. (c--f) The object can be transported between poses without collision so edges are connected. (g) The largest object cannot move between the poses of $o_t$ and $o_3$ without collision. (h) The constructed graph. }
  \label{fig:graph_construction}
\vspace{-12pt}
\end{figure*}

Fig.~\ref{fig:graph_construction} describes a full example. We omit the walls around the objects, but the objects still can move inside the boundary formed by the walls like Fig.~\ref{fig:idea}. In Fig.~\ref{fig:graph1}, nodes are generated from $O$ and the largest object ($o_2$ in dark red) is dilated by the robot size and a safety margin for conservative collision checking (gray ring). In Fig.~\ref{fig:graph2}, it is checked if the largest object can move between $(x_1, y_1)$ and $(x_2, y_2)$ (the poses of $o_1$ and $o_2$, respectively). Since there is no collision-free path, nodes $v_1$ and $v_2$ are not connected. In Figs.~\ref{fig:graph3}--\ref{fig:graph6}, the object can be transported between poses without collision so nodes are connected. The largest object collides with others if it is moved between the poses of $o_t$ and $o_3$ (Fig.~\ref{fig:graph7}) so no edge is added to $E$. Finally, the graph shown in Fig.~\ref{fig:graph_final} is constructed.  Any object in $O$ can move between two poses if a path exists between the two corresponding nodes and the objects in the path are cleared. For example, the end-effector can retrieve $o_t$ if $o_3$ and $o_1$ are sequentially removed.

The graph construction is formally described in Alg.~\ref{alg:graph}, which constructs an unweighted and undirected graph $G(V, E)$ from $O$ where $V$ and $E$ are the sets of nodes and edges, respectively. Nodes represent the objects in $O$ so $V = \{v_1, \cdots, v_N, v_t\}$ where $v_t$ is the target node (line 2). For every pair of nodes $v_i$ and $v_j$ where $i \neq j$, an edge $(i, j) \in E$ is connected if any object grasped by the end-effector can move between the poses of $o_i$ and $o_j$, which are $(x_i, y_i)$ and $(x_j, y_j)$, without a collision (lines 5--11). We use $r_g = r_{\mbox{\scriptsize max}} + r_r + r_s$ for collision checking where $r_{\mbox{\scriptsize max}}$ is the radius of the largest object, $r_r$ is the end-effector size $r_r$, and $r_s$ is the safety margin (line 4). 

For collision checking, we employ the modified VFH+ proposed in~\cite{lee2019efficient_arxiv}. It finds obstacle-free angles (directions) around an object such that the directions allow the object grasped by the end-effector to pass without collision. Specifically, if there is a collision-free pathway for the object through other objects whose breadth is greater than $2 \cdot r_g$ (notice that $r_g$ is a radius), the modified VFH+ returns a histogram where the angles toward the pathway have zero magnitude. If at least one zero-magnitude angle exists between two poses, an edge connects the nodes representing the poses. One may replace the collision checker by any available ones. 

\vspace{-5pt}
\begin{algorithm}
\caption{\textsc{GenGraph}}
\label{alg:graph}
{\footnotesize
\vspace{.01in}
\textbf{Input:}
geometry of $O$, robot size $r_r$, safety margin $r_s$\\
\textbf{Output:}
an unweighted undirected graph $G = (V, E)$\\
 
\vspace{-.1in}
1 \ \ $N = |O| - 1$ \codecomment{$N$ counts obstcales only}\\
2 \ \ $V = \{v_1, \cdots, v_N, v_t\}$ \codecomment{nodes representing all objects}\\
3 \ \ $E = \emptyset$ \\
4 \ \ $r_g = r_{\mbox{\scriptsize max}} + r_r + r_s$\codecomment{$r_{\mbox{\scriptsize max}}$ is the radius of the largest object}\\
5 \ \ \textbf{for each} $v_i \in V$ \\
6 \ \ \quad $V^\prime = V \setminus v_i$ \codecomment{no self-loop considered so $i \neq j$} \\ 
7 \ \ \quad \textbf{for each} $v_j \in V^\prime$ \\
8 \ \ \quad \quad \textbf{if} $\sim$\textsc{isCollision}$(r_g, o_i, o_j)$ \codecomment{true if there is a collision for moving the largest object between $(x_i, y_i)$ and $(x_j, y_j)$} \\
9 \ \ \quad \quad \quad $E \leftarrow E \cup (i, j)$ \codecomment{add an edge to $E$} \\
10 \ \quad \textbf{end for}\\
11 \ \textbf{end for}\\
12 \ \textbf{return} $G(V, E)$
}\end{algorithm}
\vspace{-10pt}

\smallskip
\noindent \textit{Path finding: }
In the graph $G$ from Alg.~\ref{alg:graph}, we want to find a path $V_R$, which is a sequence of nodes corresponding to the objects in $O_R$. Once all the objects in $O_R$ are removed, the end-effector can retrieve $o_t$. 

Let $V_A \in V$ be the set of nodes representing the set of accessible objects $O_A$ (\textit{accessible nodes}). Since we aim to minimize $k$ which is the number of objects to be removed, we find a minimum-hop (i.e., shortest) path from $v_i \in V_A$ to $v_t$. We use Breadth first search (BFS)~\cite{cormen2009introduction}, which is optimal and complete, to find the min-hop path between two nodes in an unweighted graph. If $|V_A| > 1$, there could be multiple paths that have the same number of hops from different starting nodes in $V_A$. 
Thus, we compute a min-hop path for each node in $V_A$. Among all the min-hop paths which may have different hops, the minimum one is chosen. If there are multiple min-hop paths, we compare their Euclidean distances to break the ties. 

\begin{figure}[h]
    \vspace{-5pt}
    \centering
    \captionsetup{skip=0pt}
   \begin{subfigure}{0.175\textwidth}
   \centering
   \captionsetup{skip=0pt}
	\includegraphics[width=0.6\textwidth]{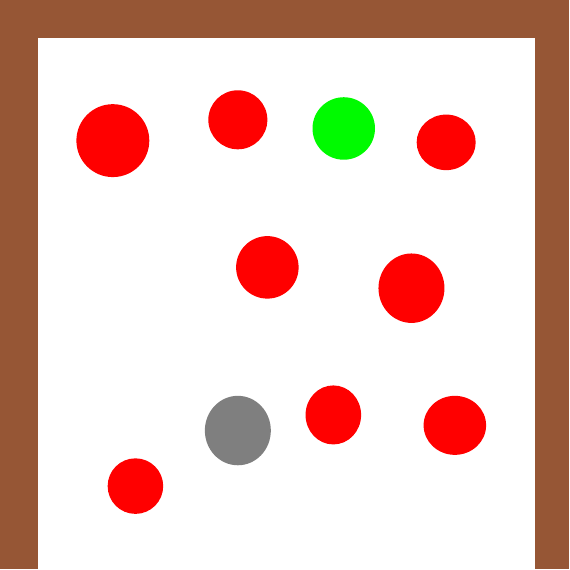}
    \caption{An object configuration}
    \label{fig:test_config}
  \end{subfigure}
  \begin{subfigure}{0.3\textwidth}
  \captionsetup{skip=0pt}
  \centering
	\includegraphics[width=0.8\textwidth]{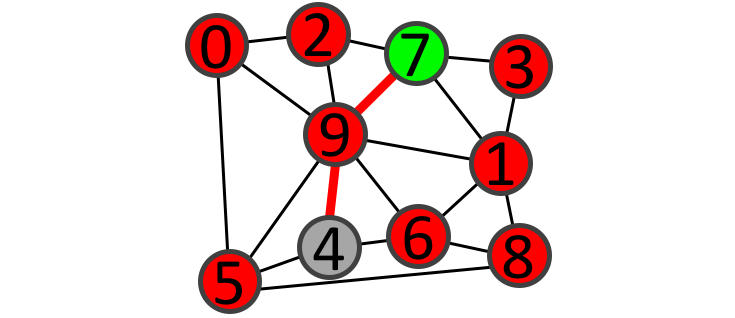}
    \caption{The graph from the configuration}
    \label{fig:test_graph}
  \end{subfigure}
    \caption{An example result from Alg.~\ref{alg:alg1}. (a) The target (green) is surrounded by obstacles (red). The gray circle is the first obstacle to be removed. (b) The red bold edges show the min-hop path.}
  \label{fig:test}
  \vspace{-15pt}
\end{figure}

\smallskip
\noindent \textit{Full algorithm: }
The full algorithm is described in Alg.~\ref{alg:alg1}. The set $V_A$ is generated from $O_A$ (line 2). A graph is constructed using Alg.~\ref{alg:graph} (line 3). For each node $v_i \in V_A$, the path with the minimum hop $k$ between $v_i$ and $v_t$ is found (line 5) using BFS. If there are multiple paths with the same $k$, the one with the shortest Euclidean distance $d$ is chosen. If the length (i.e., the number of hops) of the path $k$ is shorter than the paths stored previously, the path is updated with the new min-hop path (lines 6--9). If the path length is the same with the previously stored path but the distance is shorter, the new path replaces $P^*$ (lines 10--12). After the loop, the path without $v_t$ is the sequence of nodes $V_R$ representing the shortest path from one of the accessible object to the target. Finally, $V_R$ is converted to the sequence of objects $O_R$ (line 16).
In Fig.~\ref{fig:test}, we show an example result from Alg.~\ref{alg:alg1}. The boundary in Fig.~\ref{fig:test_config} represents walls. The green, red, and gray circles represent the target, obstacles, and the first obstacle to be removed, respectively. Fig.~\ref{fig:test_graph} shows the constructed graph. The red bold edges indicate the path $V_R = \{v_4, v_9\}$ meaning that Nodes 4 and 9 should be removed sequentially to retrieve the target.

On the other hand, graph construction does not consider the kinematic constraints of the manipulator. Only the size of the objects and the end-effector is considered for collision checking. Thus, the robot arm may collide with objects even though the end-effector grasping an object does not collide. We propose a method for dynamic replanning to incorporate the kinematic constraints. This replanning also can deal with objects that are completely occluded and appear after some front objects are removed. 

\vspace{-5pt}
\begin{algorithm}
\caption{\textsc{StaticPlanner}}
\label{alg:alg1}
{\footnotesize
\vspace{.01in}
\textbf{Input:}
geometry of $O$ and $O_A$, robot size $r_r$, safety margin $r_s$\\
\textbf{Output:}
a sequence of obstacles to be relocated $O_R$\\
 
\vspace{-.1in}
1 \ \ $P^* = \emptyset$, $k^* = \infty$, $d^* = \infty$\\
2 \ \ $V_A = \{v_i | i \mbox{ is the index of an object in } O_A\}$\codecomment{accessible nodes} \\
3 \ \ $G(V, E) = \mbox{\textsc{GenGraph}}(O, r_r, r_s)$ \codecomment{Alg.~\ref{alg:graph}}\\
4 \ \ \textbf{for each} $v_i \in V_A$ \codecomment{repeat for each accessible node}\\
5 \ \ \quad $(P, k, d) = \mbox{\textsc{MinHopPath}}(G, v_t, v_i)$ \codecomment{find a min-hop path $P$ with $k$ nodes with the Euclidean distance $d$}\\ 
6 \ \ \quad \textbf{if} $k < k^*$ \codecomment{if the path length is the shortest, update the path, length, and distance}\\
7 \ \ \quad \quad $P^* = P$ \\
8 \ \ \quad \quad $k^* = k$ \\
9 \ \ \quad \quad $d^* = d$ \\
10 \ \quad \textbf{else if} $k = k^*$ and $d < d^*$ \codecomment{if the path length is the same with $P^*$ but the Euclidean distance is shorter, update the path and distance}\\
11 \ \quad \quad $P^* = P$ \\
12 \ \quad \quad $d^* = d$ \\
13 \ \quad \textbf{end if}\\
14 \ \textbf{end for}\\
15 \ $V_R = P^* \setminus v_t$ \codecomment{exclude the target from the path}\\
16 \ $O_R = \{o_i | i \mbox{ is the index of a node in } V_R\}$\\
17 \ \textbf{return} $O_R$
}
\end{algorithm}
\vspace{-10pt}

\subsection{Case I: Dynamic replanning}
\label{sec:dynamic}
\vspace{-2pt}
Alg.~\ref{alg:alg1} would work suboptimally in dynamic situations if i) completely hidden objects are revealed while the initial plan is executed or ii) the initially computed plan is not executable owing to the robot kinematic constraints. We propose an online replanning algorithm that revises an initial relocation plan to deal with dynamic events occurring at run-time.

\smallskip
\noindent \textit{Dealing with updated object configurations: }
If hidden objects appear dynamically (Case I in Fig.~\ref{fig:uncertain}), we assume that the new objects can be added to $O$ and their geometry can be obtained. Accordingly, $O_A$ is also updated. If new objects are found, replanning follows by running Alg.~\ref{alg:alg1} with the updated configuration. If a new min-hop path is found, the previous plan is replaced. This procedure is described in lines 3--6 in Alg.~\ref{alg:alg2}.

\smallskip
\noindent \textit{Incorporating robot kinematic constraints: }
Once $O_R$ is computed, the robot needs to compute trajectories for the whole manipulator (including the end-effector and the robot arm both) using any off-the-shelf motion planner. However, a motion planner would fail to generate a trajectory satisfying the kinematic constraints of the robot. Then the robot removes additional obstacles to comply with the constraints. Suppose that the path shown in Fig.~\ref{fig:test} is not executable because the robot arm (not the end-effector) collide with $o_6$ when the end-effector approaches to $o_9$ after removing $o_4$. Then the robot can modify the plan to remove $o_6$ before removing $o_9$. This procedure is described in lines 10--14 in Alg.~\ref{alg:alg2}. Note that \textsc{ColObj} in line~10 can be implemented using collision checking libraries like~\cite{pan2012fcl}.

\vspace{-7pt}
\begin{algorithm}
\caption{\textsc{DynamicPlanner}}
\label{alg:alg2}
{\footnotesize
\vspace{.01in}
\textbf{Input:}
$O$, $O_A$, robot kinematics $X$, robot size $r_r$, safety margin $r_s$\\
\textbf{Output:}
Done\\

\vspace{-.1in}
1 \ \ $O_R = $\,\textsc{StaticPlanner}$(O, O_A, r_r, r_s)$\codecomment{compute the initial plan}\\
2 \ \ \textbf{while} $o_t \in O$ is not grasped\codecomment{remove each object until done}\\
\codecommentline{Lines 3--6: dealing with updated object configurations}\\
3 \ \ \quad Update $O$ and $O_A$ with new sensing data\\
4 \ \ \quad \textbf{if} $O$ or $O_A$ is changed\\
5 \ \ \quad \quad $O_R = $\,\textsc{StaticPlanner}$(O, O_A, r_r, r_s)$\\\codecomment{run Alg.~\ref{alg:alg1} again with the updated configuration}\\
6 \ \ \quad \textbf{end if}\\
7 \ \ \quad \textbf{if} the first object in $O_R$ is graspable\\
8 \ \ \quad \quad \textsc{Dequeue}$(O_R)$\codecomment{remove the object from the queue and the scene}\\
9 \ \ \quad \textbf{else} \\
\codecommentline{Lines 10--14: incorporating robot kinematic constraints}\\
10 \ \quad \quad $O_C = $\,\textsc{ColObj}$(O, O_R, X)$\codecomment{find objects colliding with the arm}\\
11 \ \quad \quad \textbf{for each} $o_c \in O_C$\\
12 \ \quad \quad \quad $o_t = o_c$\codecomment{$o_c$ becomes a temporary target}\\
13 \ \quad \quad \quad \textsc{DynamicPlanner}$(O, O_A, X, r_r, r_s)$\\\codecomment{recursively remove all colliding objects}\\
14 \ \quad \quad \textbf{end for}\\
15 \ \quad \textbf{end if}\\
16 \ \textbf{end while}\\
17 \ \textbf{return} Done
}\end{algorithm}
\vspace{-15pt}

\subsection{Case II: Search for the target}
\label{sec:uncertain}
\vspace{-2pt}
Occlusions cause uncertainties in recognizing objects so the geometry of $O$ can be partially known and the target may not be detected (Case II in Fig.~\ref{fig:uncertain}). The robot is tasked with target search in order to retrieve the target. If the target is detected, Alg.~\ref{alg:alg1} or Alg.~\ref{alg:alg2} generates a relocation plan. 

In~\cite{dogar2014object}, an object is chosen to be removed such that  the volume revealed after the removal is maximized. We develop three simple search strategies including the one similar to \cite{dogar2014object}\footnote{We do not consider the overlaps in the occluded volumes of objects for simplicity. Thus, the volume is calculated for each object independently.} (which we call \textit{Volume} strategy). Other two strategies are based on the Euclidean distance between the end-effector and objects. \textit{Closest} and \textit{Farthest} remove the object with the shortest and longest distance from the robot, respectively. The quantity $m_i$ in lines~4 and 6 of Alg.~\ref{alg:alg3} is determined depending on the strategy used.

\vspace{-7pt}
\begin{algorithm}
\caption{\textsc{UncertainPlanner}}
\label{alg:alg3}
{\footnotesize
\vspace{.01in}
\textbf{Input:}
geometry of $O$ and $O_A$, robot size $r_r$, safety margin $r_s$\\
\textbf{Output:}
a sequence of obstacles to be relocated $O_R$\\
 
\vspace{-.1in}
1 \ \ \textbf{while} $o_t$ is not detected\\
2 \ \ \quad Update $O$ and $O_A$\\
3 \ \ \quad \textbf{for each} $o_i \in O_A$\\
4 \ \ \quad \quad Compute the metric $m_i$\codecomment{revealed volume or distance}\\
5 \ \ \quad \textbf{end for}\\
6 \ \ \quad $o_r = \argmax_{o_i \in O_A} m_i$\codecomment{$\argmin$ for \textit{Closest} strategy}\\
7 \ \ \quad Relocate $o_r$\\
8 \ \ \textbf{end while}\\
9 \ \  Update $O$ and $O_A$\\
10 \  $O_R =$\,\textsc{StaticPlanner}$(O, O_A, r_r, r_s)$\\\codecomment{or \textsc{DynamicPlanner} in dynamic environments}\\
11 \ \textbf{return} $O_R$\codecomment{\textit{Done} is returned if Alg.~\ref{alg:alg2} is executed in line 10}
}\end{algorithm}
\vspace{-15pt}

\subsection{Analysis of algorithms}
\label{sec:analysis}
\vspace{-2pt}
We prove time complexity and completeness of the algorithms. 

\lem \textbf{3.1.} Alg.~\ref{alg:graph} has polynomial time complexity.

\pf. Collision checking (line~8) runs for every pair of objects (lines~5--11). There are total $N(N+1)/2$ pairs. The collision checker that we used runs in $O(N(N+N)) = O(N^2)$~\cite{lee2019efficient_arxiv}. Thus, the time complexity is $O((N(N+1)/2) \times (N(N+N))) = O(N^4)$. \qed
\smallskip

\thm \textbf{3.2.} Alg.~\ref{alg:alg1} has polynomial time complexity.

\pf. The graph $G(V, E)$ has $N+1$ nodes and at most $N(N+1)/2$ edges (fully connected). BFS runs in time $O(|E| + |V|) = O(N(N+1)/2 + N+1) = O(N^2)$. In line~5, we find all min-hop paths between a pair of nodes to choose the one with the shortest Euclidean distance. Finding all min-hop paths is done by running BFS for all nodes in $V_A$, which takes $O((N^2)N)$ as $|V_A| \le N+1$. Thus, lines~4--14 runs in $O(N^4)$. By Lemma~3.1, line~3 takes $O(N^4)$ so the time complexity is $O(N^4 + N^4) = O(N^4)$. \qed
\smallskip

\thm \textbf{3.3.} Alg.~\ref{alg:alg1} is complete if $G$ is connected\footnote{A graph is connected if there is a path between every pair of nodes. A graph that is not connected has more than one nodes which are completely isolated (so has no edge).}.

\pf. First, we want to show that Alg.~\ref{alg:graph} is complete. Collision checking in line~8 of Alg.~\ref{alg:graph} is complete~\cite{lee2019efficient_arxiv}, which means that $G$ is constructed after a finite number of iterations.
By definition of connected graphs, a path exists from one of the accessible nodes to the target. BFS is complete~\cite{cormen2009introduction} so finds the shortest path. Thus, Alg.~\ref{alg:alg1} always returns a path which is the shortest. \qed
\smallskip

\thm \textbf{3.4.} Alg.~\ref{alg:alg2} has polynomial time complexity. 

\pf. Suppose that $M$ hidden objects are added to $O$ and $L$ objects need to be removed owing to robot kinematic constraints. The while loop (lines~2--16) runs at most $N+M-L$ times until only the target remains. Inside the loop, Alg.~\ref{alg:alg1} in line~5 runs with at most $N+M$ objects if all $M$ objects are revealed. Alg.~\ref{alg:alg2} runs at most $L$ times if an object collides in every step. Using Theorem 3.1, the time complexity of Alg.~\ref{alg:alg2} is $O\big((N+M-L) \times \{(N+M)^4 + L((N+M-L)(N+M)^4)\}\big)$ (notice that Alg.~\ref{alg:alg2} is being called recursively at most $L$ times). Since objects colliding with the robot cannot exceed the number of total objects excluding the target, $0 \le L \le N+M$. Thus, the time complexity is $O\big((N+M) \times \{(N+M)^4 + (\frac{N+M}{2})^2(N+M)^4\}\big) = O\big((N+M) \times (N+M)^6\big) = O((N+M)^7)$.\qed
\smallskip

\thm \textbf{3.5.} Alg.~\ref{alg:alg2} is complete if $M$ is finite.

\pf. The proof is done by contradiction. Suppose that $M$ is finite but Alg.~\ref{alg:alg2} cannot find a solution. Since there is at least an accessible object ($O_A \neq \emptyset$), $O_R$ must be returned in line~1 by Theorem 3.3. The while loop (lines 2--16) removes at least one object in each iteration and terminates after $N+M$ iterations at most. Thus, Alg.~\ref{alg:alg2} terminates and return the output in a finite time. The supposition is false. \qed
\smallskip

\thm \textbf{3.6.} Alg.~\ref{alg:alg3} runs in polynomial time. 

\pf. Computing $m_i$ takes a constant time with a closed-form solution of the volume or distance. After computing the metric for all known objects (at most $N$), one object should be chosen using a maximum or a minimum operation running in $O(N)$. Thus, an $O(N^2)$ operation is needed to decide what to relocate. Once the target is detected, Alg.~\ref{alg:alg1} or Alg.~\ref{alg:alg2} runs. The time complexity of Alg.~\ref{alg:alg3} is $O(N^2 + N^4) = O(N^4)$ with Alg.~\ref{alg:alg1} or $O((N+M)^2 + (N+M)^7) = O((N+M)^7)$ with Alg.~\ref{alg:alg2}. \qed
\smallskip

Alg.~\ref{alg:alg3} does not necessarily need a connected graph. Although $G$ is not connected initially, it must be connected as obstacles are removed. In the worst case, only the target node remains. A singleton graph is connected by definition. 

\section{Experiments}
\label{sec:exp}
\vspace{-2pt}
In this section, we measure the planning time of the algorithms and the total running time in a simulated environment and using a physical robot integrated with a vision system. 

\subsection{Scenarios}
\begin{figure}[h]
\vspace{-5pt}
    \centering
    \captionsetup{skip=0pt}
   \begin{subfigure}{0.215\textwidth}
   \captionsetup{skip=0pt}
	\includegraphics[width=\textwidth]{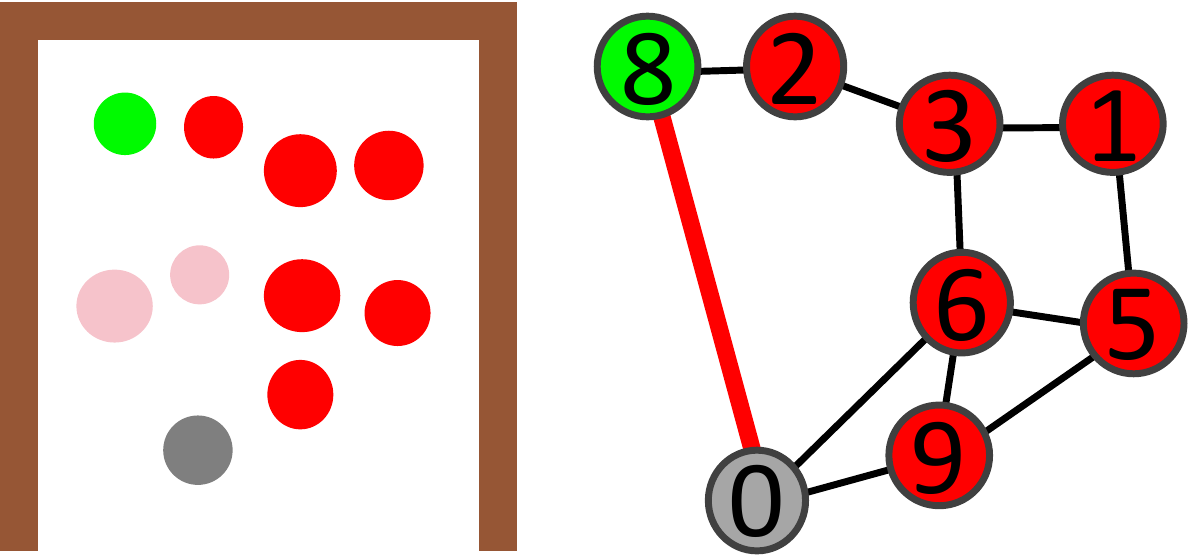}
    \caption{The initial path with two hidden objects (pink)}
    \label{fig:dynamic_ex_a}
  \end{subfigure}\quad
  \begin{subfigure}{0.22\textwidth}
  \captionsetup{skip=0pt}
	\includegraphics[width=\textwidth]{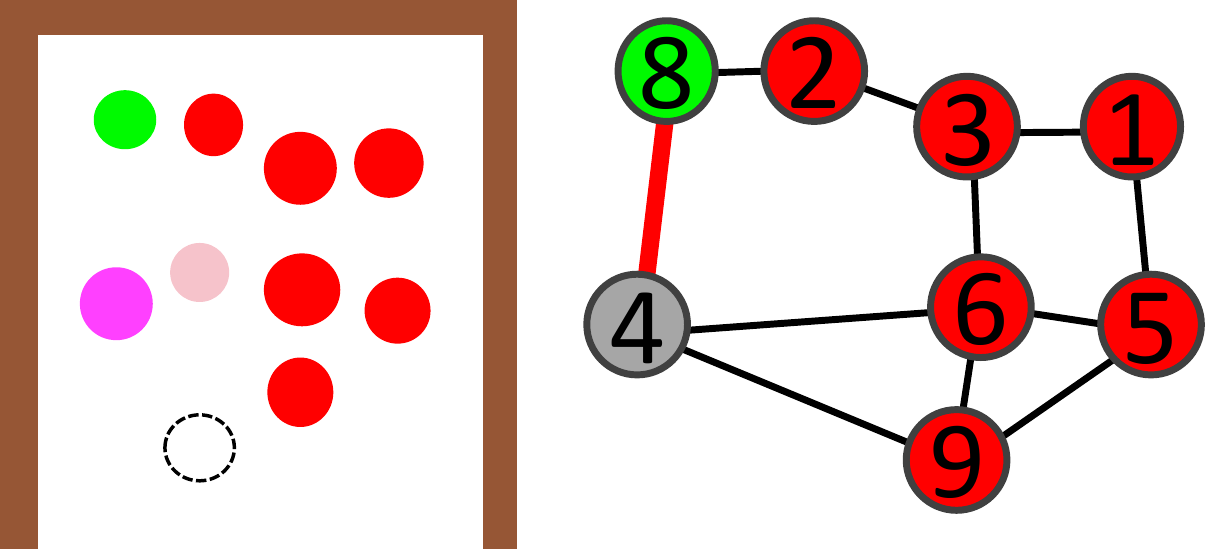}
    \caption{An object revealed (magenta)}
    \label{fig:dynamic_ex_b}
  \end{subfigure}
  \begin{subfigure}{0.22\textwidth}
  \captionsetup{skip=0pt}
	\includegraphics[width=\textwidth]{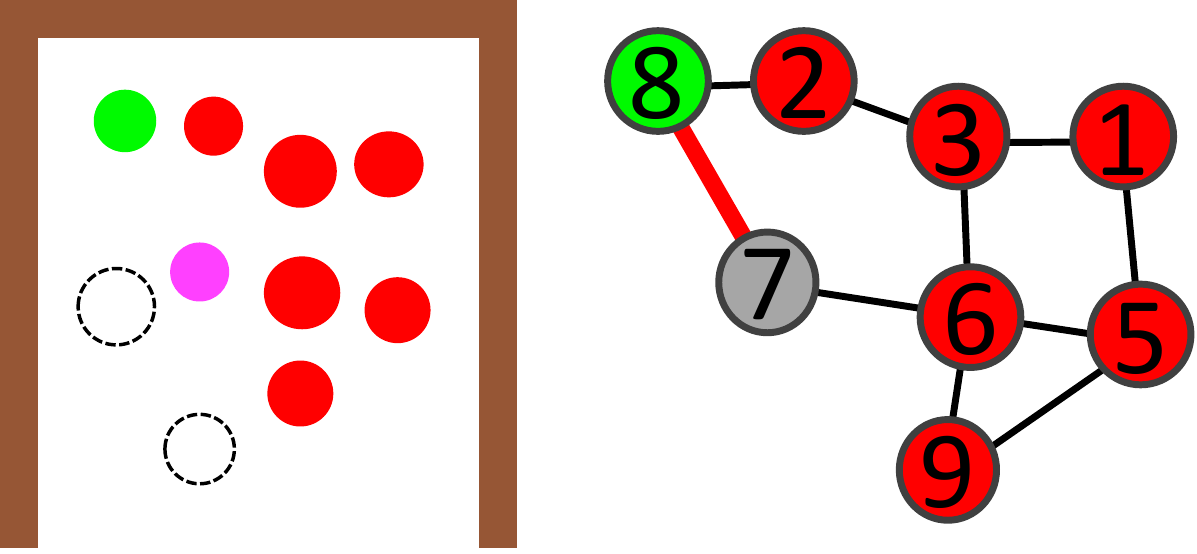}
    \caption{Another object revealed}
    \label{fig:dynamic_ex_c}
  \end{subfigure}\quad
  \begin{subfigure}{0.22\textwidth}
  \captionsetup{skip=0pt}
	\includegraphics[width=\textwidth]{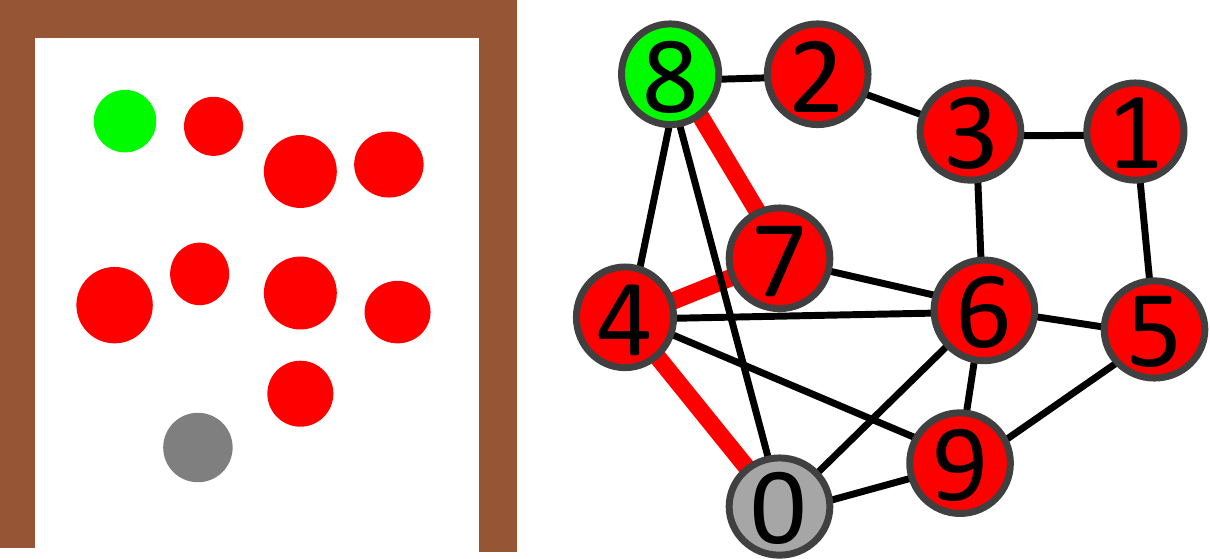}
    \caption{The final path}
    \label{fig:dynamic_ex_final}
  \end{subfigure}
    \caption{An example of Scenario 2. (a) The two pink objects are hidden. (b) After $o_0$ is removed, $o_4$ (in magenta) occurs so a new path is computed. (c) After $o_4$ is removed, $o_7$ is revealed. (d) The final path is shown.}
  \label{fig:dynamic_ex}
  \vspace{-12pt}
\end{figure}

\begin{figure}[h]
\vspace{-5pt}
    \centering
    \captionsetup{skip=0pt}
   \begin{subfigure}{0.273\textwidth}
   \captionsetup{skip=0pt}
	\includegraphics[width=\textwidth]{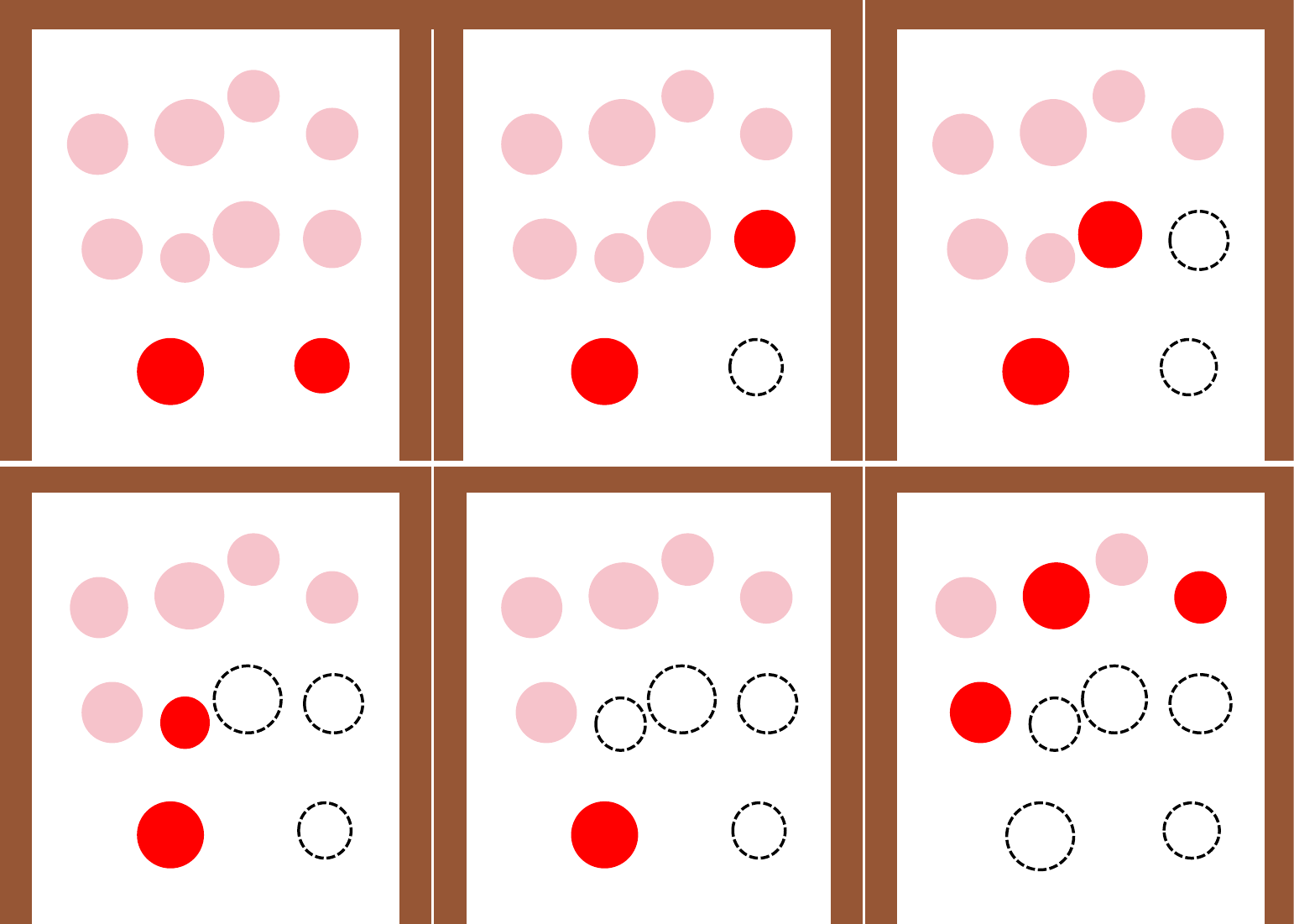}
    \caption{Search (from top left to bottom right)}
    \label{fig:uncertain_ex_progress}
  \end{subfigure}
  \begin{subfigure}{0.20\textwidth}
  \captionsetup{skip=0pt}
	\includegraphics[width=\textwidth]{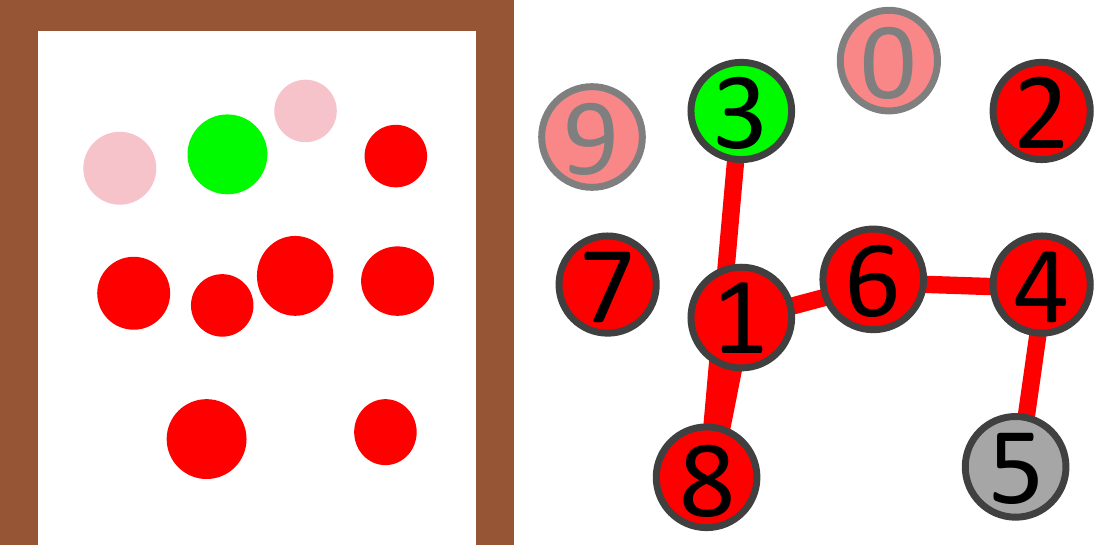}
    \caption{The final configuration (left) and the path nodes in the graph (right)}
    \label{fig:uncertain_ex_final}
  \end{subfigure}
    \caption{An example of Scenario 3 (with \textit{Farthest} strategy). (a) An object is chosen to be removed in each step. Some new objects are discovered after a relocation. (b) Finally, the target is detected.}
  \label{fig:uncertain_ex}
  \vspace{-12pt}
\end{figure}

\vspace{-2pt}
We consider three scenarios summarized in Table~\ref{tab:scenario}. The known geometry of $O$ means that full geometric information (e.g., sizes, poses) of objects including the target are known. The target is detected if the geometry of the target is known. Scenario 1 considers the detected target and known geometry (Base case). An example is shown in Fig.~\ref{fig:test}. In Scenario 2, the target is detected, but 20\% of the total objects are hidden. Fig.~\ref{fig:dynamic_ex} shows an example where hidden objects are revealed if they are recognized by the robot.\footnote{We do not use an actual recognition algorithm determining visibility of objects as it is out of scope of this work. Instead, we simplify the process by imposing a consistent criteria: a hidden object is revealed if it is accessible to the robot.} Scenario 3 is with an undetected target and partially known geometry. Fig.~\ref{fig:uncertain_ex} shows several steps until the target is recognized. In this example, we assume that an accessible object is not occluded so visible to the robot. 

\begin{table}[h!]
\vspace{-7pt}
\captionsetup{skip=0pt}
\caption{Scenarios used in the experiments}
\label{tab:scenario}
\centering
\scalebox{0.95}{
\begin{tabular}{|c||c|c|c|}
\hline
Scenario & Target & Geometry of $O$ & Used algorithm\\
\hline
1 (Base case) & Detected & Known & Alg.~\ref{alg:alg1}\\
\hline
2 (Case I) & Detected & Partially known & Alg.~\ref{alg:alg2} \\
\hline
3 (Case II) & Undetected & Partially known & Alg.~\ref{alg:alg3} \\
\hline
\end{tabular}}
\vspace{-12pt}
\end{table}


\subsection{Computation time for planning}
\begin{figure}[h]
\vspace{-7pt}
\captionsetup{skip=0pt}
    \centering
	\includegraphics[scale=0.4]{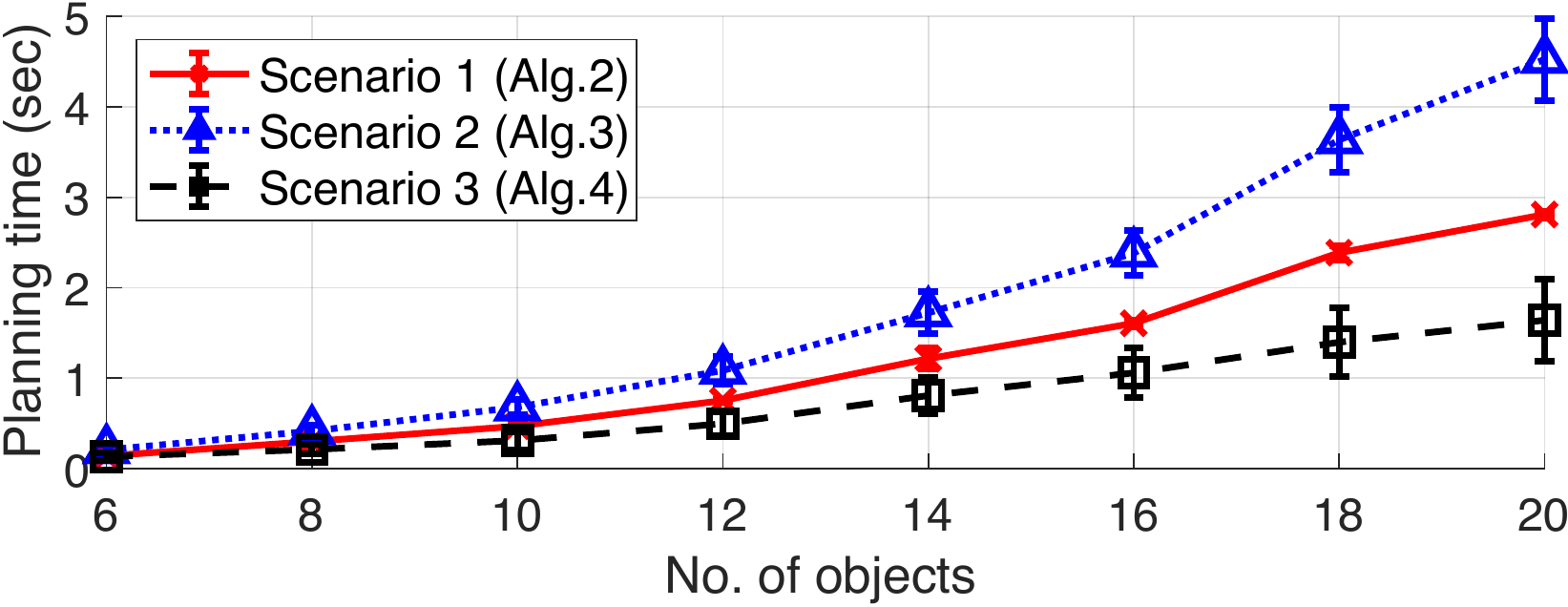}
    \caption{Computation time (20 repetitions for each instance size).}
    \label{fig:perf}
    \vspace{-15pt}
\end{figure}

\vspace{-2pt}
We test the algorithms with random instances to measure the computation time. We randomly generate 20 instances for each instance size where the number of objects including the target is from 6 to 20 incremented by 2. The system is with Intel Core i7 2.7GHz with 16G RAM and Python 3.7. The result is shown in Fig.~\ref{fig:perf} and Table~\ref{tab:perf}. In Scenario 1, Alg.~\ref{alg:alg1} can compute a relocation plan for 20 objects within 3\,sec. In other words, the robot needs less than 3 seconds until it starts executing a relocation plan for 20 objects. Considering the relatively long execution time of a relocation plan which takes few minutes, the planning time is very short. For a moderate-sized instance (e.g., 10 objects), Alg.~\ref{alg:alg1} can produce a solution very quickly (less than 0.5\,sec). 
In Scenarios 2 and 3, the algorithms update the geometric information and replan in each iteration after each object is relocated. We measure the average computation time per iteration, which the robot needs before manipulating an object in each iteration. For 20 objects, Algs.~\ref{alg:alg2}--\ref{alg:alg3} take 4.5 and 1.6\,sec, respectively. The planning time using Alg.~\ref{alg:alg2} is relatively longer due to its high order of time complexity but is not prohibitively long to be used practically. For example, it takes less than 0.7\,sec with 10 objects.

\begin{table*}
\captionsetup{skip=0pt}
\caption{The planning time (sec) and standard deviation (in parentheses) of the proposed algorithms in the three scenarios  (20 repetitions)}
\label{tab:perf}
\centering
\scalebox{0.91}{
\begin{tabular}{|c|c|c|c|c|c|c|c|c|}
\hline
\multirow{2}{*}{Scenario ID} & \multicolumn{8}{c|}{Number of objects} \\
\cline{2-9}
  & 6 & 8 & 10 & 12 & 14 & 16 & 18 & 20\\
\hline
1 (Alg.~\ref{alg:alg1}) & 0.01 (0.02) & 0.30 (0.03) & 0.47 (0.02) & 0.75 (0.01) & 1.22 (0.12) & 1.61 (0.04) & 2.39 (0.08) & 2.81 (0.04)  \\
\hline
2 (Alg.~\ref{alg:alg2}) & 0.21 (0.05) & 0.42 (0.06) & 0.68 (0.09) & 1.09 (0.15) & 1.72 (0.23) & 2.38 (0.25) & 3.64 (0.36) & 4.53 (0.45) \\
\hline
3 (Alg.~\ref{alg:alg3}) & 0.13 (0.04) & 0.21 (0.05) & 0.31 (0.10) & 0.50 (0.13) & 0.81 (0.20) & 1.06 (0.27) & 1.40 (0.38) & 1.64 (0.46) \\
\hline
\end{tabular}}
\vspace{-5pt}
\end{table*}

\subsection{Experiments in simulated environments}
\begin{figure}[h]
\vspace{-7pt}
\captionsetup{skip=1pt}
    \centering
	\includegraphics[scale=0.1]{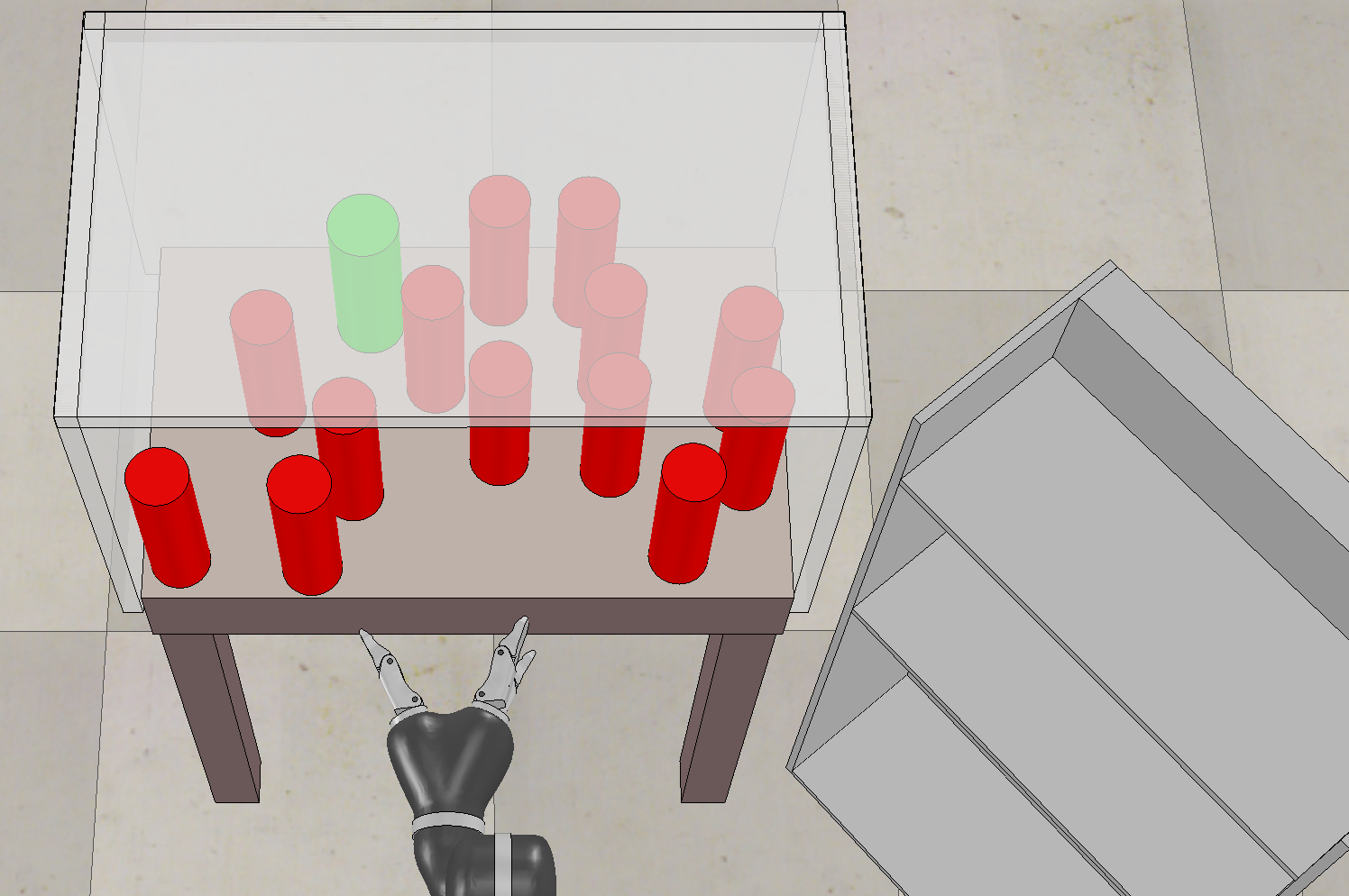}
    \caption{The simulated environment implemented in V-REP.}
    \label{fig:vrep}
    \vspace{-10pt}
\end{figure}

\vspace{-2pt}
We test the proposed algorithms in a simulated environment using a high fidelity robotic simulator V-REP~\cite{rohmer2013v} with Vortex physics engine. The system for simulation is with Intel Core i7 4.20GHz and 8GB RAM. We use a model of Kinova JACO1, which is a 6-DOF manipulator. 
Once a relocation plan is computed, the trajectory of the manipulator is generated using the Open Motion Planning Library (OMPL)~\cite{sucan2012the-open-motion-planning-library} (we use BIT-RRT~\cite{devaurs2013enhancing}). The values used for the robot size $r_r$ and safety margin $r_s$ are 3.5\,cm and 0.5\,cm, respectively.
In Scenario 1, we test instances where the numbers of objects including the target are 6, 10, 14, and 18, respectively. In other scenarios, we test instances with 10 objects. Objects are randomly populated on a 0.7\,m by 0.5\,m tabletop, whose diameters are randomly sampled from $\mathcal{U}(5, 6)$ whose unit is centimeter. We generate 10 random instances for each size where instances do not need any relocation are discarded. The robot base is fixed in front of the table as shown in Fig.~\ref{fig:vrep}. The robot place removed obstacles in the shelf next to the table. We measure the number of relocated objects and the total running time which includes planning and execution.

\begin{figure}[h]
\vspace{-5pt}
\captionsetup{skip=0pt}
    \centering
   \begin{subfigure}{0.23\textwidth}
   \centering
   \captionsetup{skip=0pt}
	\includegraphics[width=\textwidth]{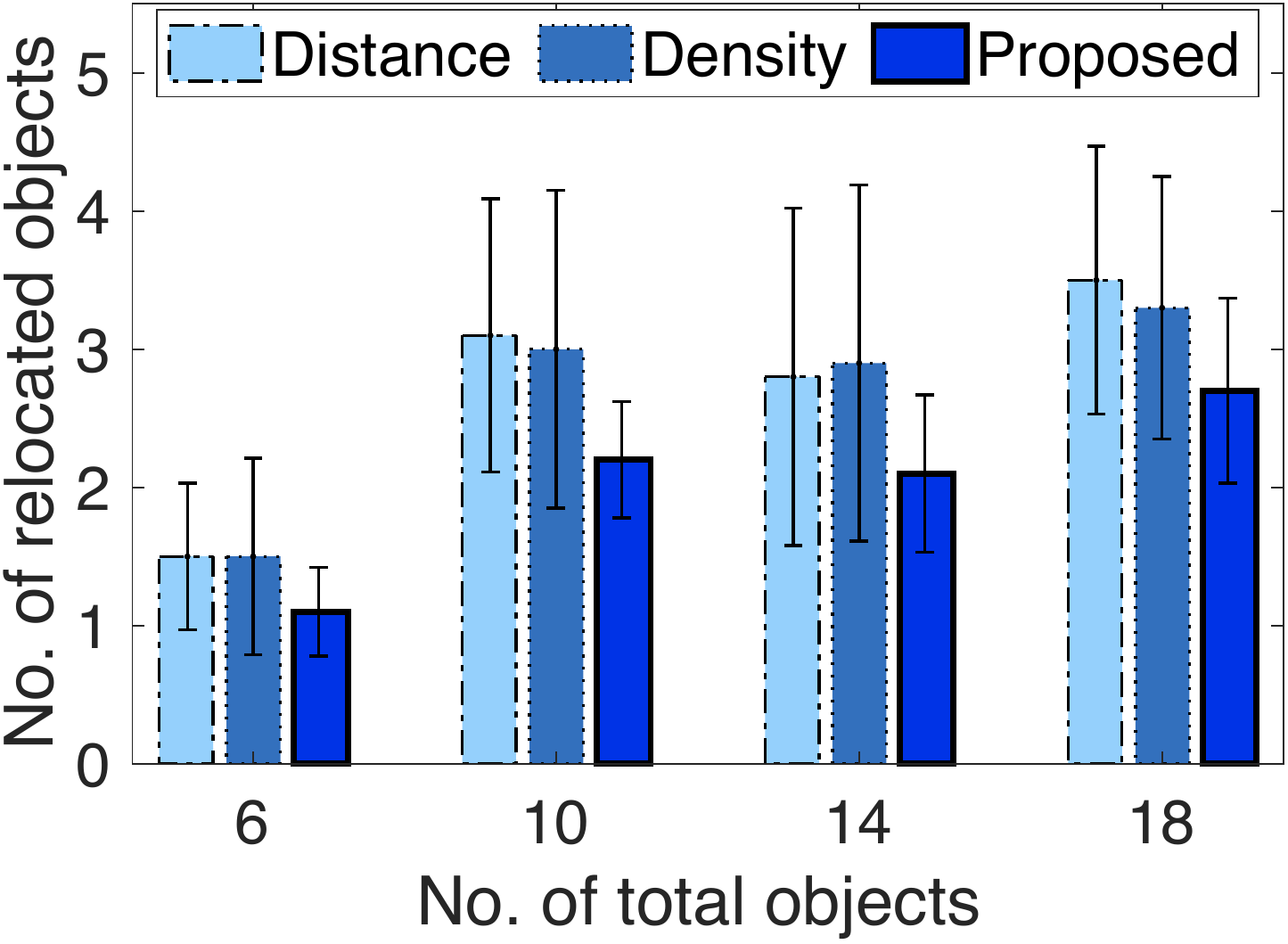}
    \caption{The number of relocated objects}
    \label{fig:sim_num}
  \end{subfigure}%
  \begin{subfigure}{0.23\textwidth}
  \centering
  \captionsetup{skip=0pt}
	\includegraphics[width=\textwidth]{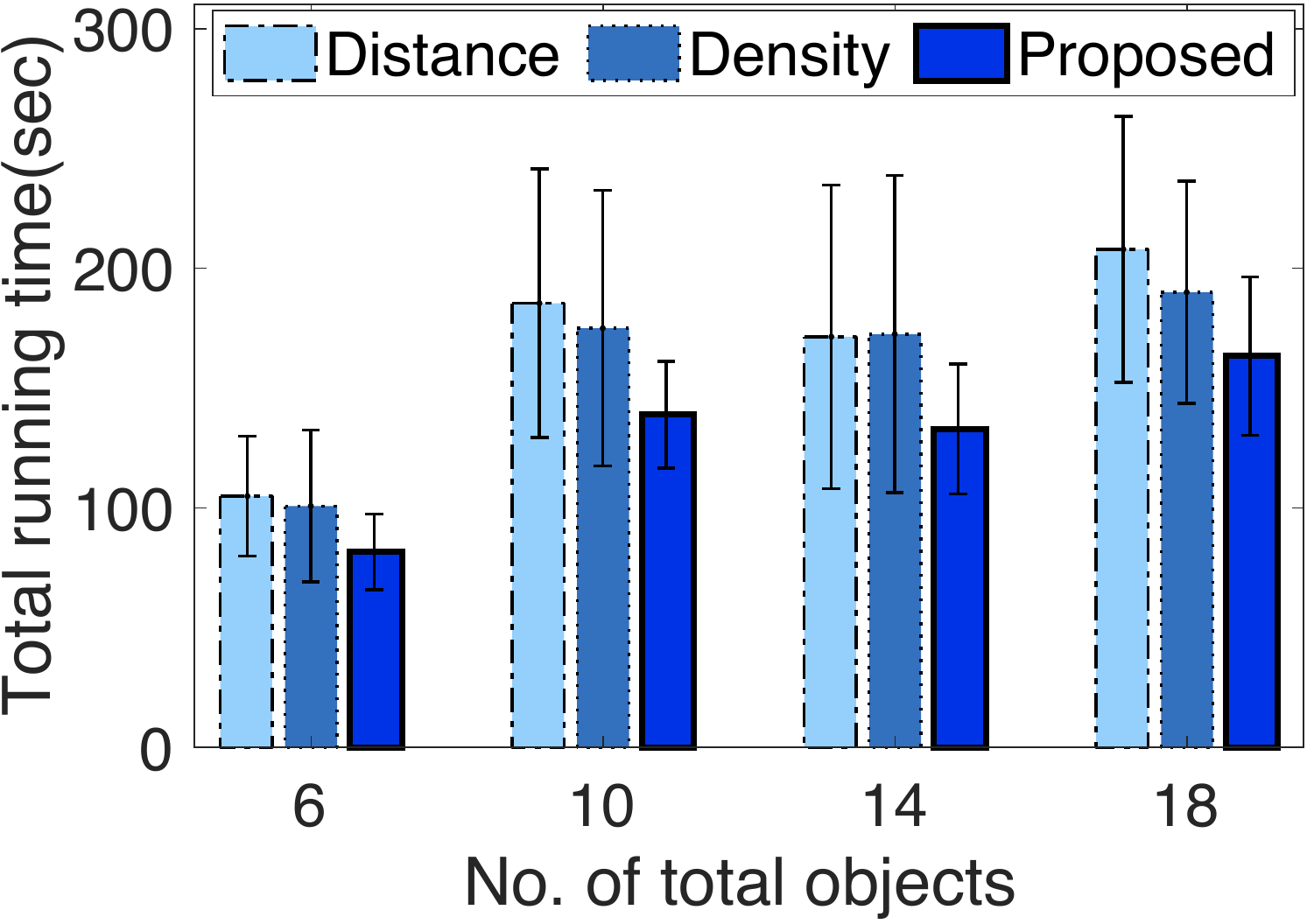}
    \caption{The total running time}
    \label{fig:sim_time}
  \end{subfigure}
    \caption{The comparison between different algorithms in Scenario 1}
  \label{fig:sim}
  \vspace{-7pt}
\end{figure}

In Scenario 1, we compare Alg.~\ref{alg:alg1} with i) a baseline method used in~\cite{dogar2012planning} which removes obstacles on the distance-optimal path of the end-effector (\textit{Distance}) and ii) the method presented in~\cite{zacharias2006bridging} which chooses an obstacle to remove greedily in the direction with the lowest density of obstacles (\textit{Density}). In Scenario 2, we compare Alg.~\ref{alg:alg2} with a replanning method based on \textit{Distance}. The robot using \textit{Distance} removes all obstacles in the shortest path including newly revealed ones (an example is shown in the accompanying video material). In Scenario 3, we compare the three strategies, \textit{Volume}, \textit{Closest}, and \textit{Farthest}. For each repetition with a random object configuration, we use the same configuration to compare different algorithms. The result is summarized in Fig.~\ref{fig:sim} and Table~\ref{tab:sim}. 

\begin{table}[h]
\captionsetup{skip=0pt}
\caption{The results of simulations in V-REP (10 repetitions). The average number of relocated objects (\#relocation) and total running time are compared with other methods.}
\label{tab:sim}
\centering
\begin{subtable}[t]{0.95\linewidth}
\centering
\scalebox{0.95}{
\begin{tabular}{|c|c|c|c|c|}
\hline
No. of & \multirow{2}{*}{Metric} & \multicolumn{3}{c|}{Method} \\
\cline{3-5}
objects & & Distance & Density & Proposed \\
\hline
\multirow{2}{*}{6} & \#relocation & 1.5 (0.53) & 1.5 (0.71) & 1.1 (0.32) \\
\cline{2-5}
& Time (sec) & 104.9 (25.0) & 100.8 (31.7) & 81.6 (15.8) \\
\hline
\multirow{2}{*}{10} &  \#relocation & 3.1 (0.99)  & 3.0 (1.15) & 2.2 (0.42) \\
\cline{2-5}
& Time (sec) & 185.4 (56.1) & 175.0 (57.5) & 138.9 (22.3) \\
\hline
\multirow{2}{*}{14} &  \#relocation & 2.8 (1.22)  & 3.0 (1.29) & 2.1 (0.57)\\
\cline{2-5}
& Time (sec) & 171.4 (63.3)  & 172.5 (66.2) & 132.9 (27.1) \\
\hline
\multirow{2}{*}{18} &  \#relocation & 3.5 (0.97) & 3.3 (0.95) &  2.7 (0.67) \\
\cline{2-5}
& Time (sec) & 207.9 (55.5)  & 190.0 (46.3) &  163.36 (33.1) \\
\hline
\end{tabular}
}
\caption{Scenario 1} 
\label{tab:sim_num}
\end{subtable}\vspace{-2pt}

\begin{subtable}[t]{0.95\linewidth}
\captionsetup{skip=0pt}
\centering
\scalebox{0.95}{
\begin{tabular}{|c|c|c|}
\hline
 Metric &  Distance & Proposed \\
\hline
\#relocation & 2.9 (1.73) & 2.1 (0.74) \\
\hline
Time (sec) & 163.2 (80.1) & 131.9 (36.8)\\
\hline
\end{tabular}
}
\caption{Scenario 2 (10 objects)} 
\label{tab:sim_scenario2}
\end{subtable}\vspace{3pt}

\begin{subtable}[t]{0.95\linewidth}
\captionsetup{skip=0pt}
\centering
\scalebox{0.95}{
\begin{tabular}{|c|c|c|c|}
\hline
 Metric & Volume & Closest & Farthest\\
\hline
\#relocation & 3.8 (2.19) & 4.4 (2.11) & 3.7 (2.17)\\
\hline
Time (sec) & 213.6 (118.2) & 239.9 (107.8) & 210.9 (111.8) \\
\hline
\end{tabular}
}
\caption{Scenario 3 (10 objects)} 
\label{tab:sim_scenario3}
\end{subtable}\vspace{-10pt}
\end{table}

\noindent \textit{Analysis}: In Scenario 1, Alg.~\ref{alg:alg1} outperforms \textit{Distance} and \textit{Density} as shown in Fig.~\ref{fig:sim}. With 10 objects, it reduces the number of relocation up to 29.0\% and the running time up to 25.1\%. 46.5\,sec are saved compared to \textit{Distance}. With other instance sizes, the reduction ranges from 18.2 to 27.6\% and 14.0 to 23.0\% in the number of relocation and the running time, respectively. The smallest reduction in running time is 19.2\,sec which is still significant (six objects). The results from Alg.~\ref{alg:alg1} have lower variances than others showing that ours produces a better solution consistently while others have the performance fluctuating depending on the randomness of the object configurations.
In Scenario 2, Alg.~\ref{alg:alg2} reduces the number of relocation 27.6\% and the running time 19.1\% (31.2\,sec) compared to \textit{Distance}. In Scenario 3, all the three strategies relocate more obstacles compared to other scenarios because the robot needs an additional number of relocation until it locates the target. The additional moves cause significantly longer running time (at most 101\,sec). \textit{Volume} and \textit{Farthest} show comparable results. \textit{Farthest} is easier to implement since it simply computes the distance to the objects but does not need to compute the occluded volume which could be complicated if objects have irregular shapes. \textit{Closest} shows a clear difference in the performance indicating that going deeper into the workspace as much as possible is more effective than clearing the foremost ones.

\subsection{Experiments with a physical robot and a vision system}
\label{sec:jaco}
\begin{figure}[h]
\vspace{-5pt}
\captionsetup{skip=0pt}
    \centering
   \begin{subfigure}{0.24\textwidth}
   \centering
   \captionsetup{skip=0pt}
	\includegraphics[width=\textwidth]{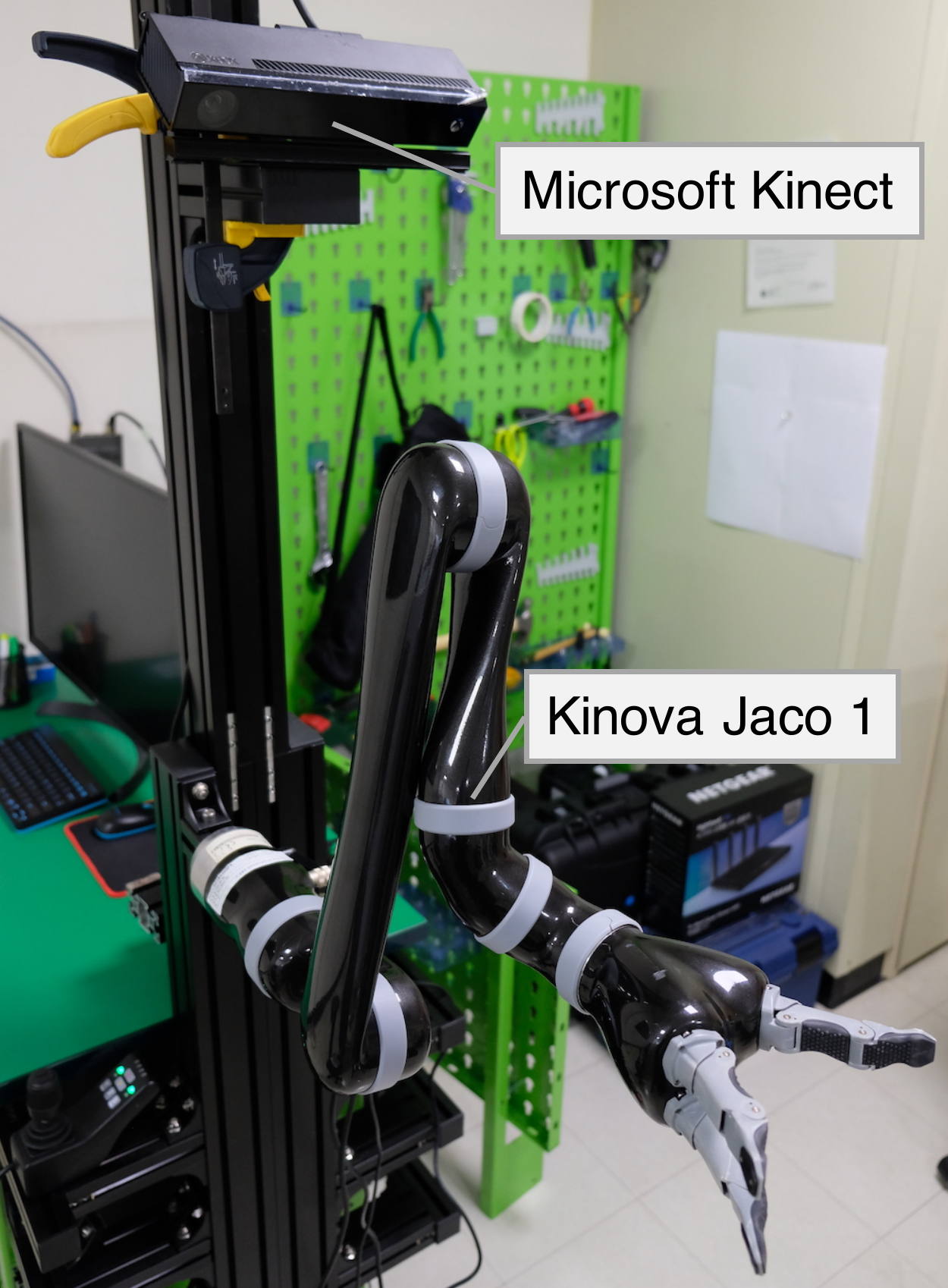}
    \caption{Kinova Jaco 1 and Kinect V2}
    \label{fig:jaco}
  \end{subfigure}
  \begin{subfigure}{0.25\textwidth}
  \centering
  \captionsetup{skip=0pt}
	\includegraphics[width=0.97\textwidth]{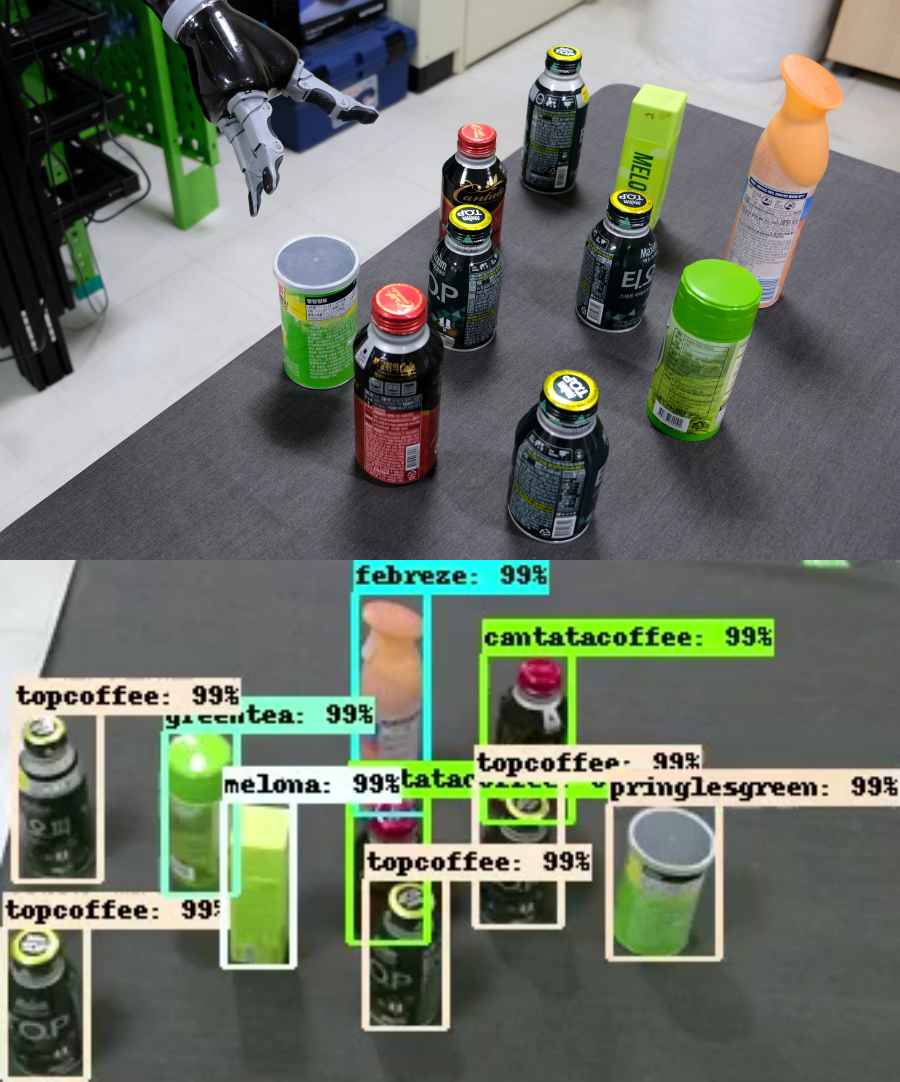}
    \caption{(Top) An object configuration (Bottom) Object detection using deep neural networks~\cite{ren2017faster}}
    \label{fig:vision}
  \end{subfigure}
    \caption{The system with an example configuration}
  \label{fig:robot}
  \vspace{-10pt}
\end{figure}

\vspace{-2pt}
We validate Alg.~\ref{alg:alg1} using a system integrating a physical robot and a vision. Other algorithms are not experimented owing to the space limit. Fig.~\ref{fig:robot} shows the system and the environment used in the experiment. We use a Kinova Jaco 1 with a fixed base. An RGB-D sensor (Kinect V2) is installed above the manipulator where the whole workspace of the robot can be captured. We implement \textit{Faster R-CNN}~\cite{ren2017faster} to detect objects where object information is displayed in the bounding boxes shown in Fig.~\ref{fig:vision}. From the 3D point cloud data, we compute the geometry of the objects (e.g., the grasp centers and sizes). Once a relocation plan is computed, each object in the plan is grasped by following a path generated by RRT where the motions are computed by solving inverse kinematics of the robot. 
We generate 10 random instances with 10 objects where instances do not need any relocation are discarded. For each instance, we compare Alg.~\ref{alg:alg1} and \textit{Distance} (a comparison is shown in the accompanying video material). In average, Alg.~\ref{alg:alg1} and \textit{Distance} relocate 1.4 and 2.1 objects, respectively. Their standard deviations are 0.52 and 0.57. The average total running time of Alg.~\ref{alg:alg1} is 174.5\,sec ($\sigma = 48.9$) while \textit{Distance} takes 229.9\,sec ($\sigma =49.5$). The reduction of the number of relocation and running time are 33.3\% and 24.1\%, respectively. 


\section{Conclusion}
\vspace{-2pt}
In this work, we study the problem of retrieving objects from clutter without collisions. Our objective is to minimize the number of objects to be relocated to generate a collision-free path for the end-effector of a robotic manipulator so as to reduce the total running time to retrieve the target object. In addition to known environments, we consider partially known environments incurred by occlusions. We develop polynomial-time and complete algorithms. The results from extensive experiments show that our methods reduce the entire running time significantly, compared to a baseline method. The experiment with a physical robot and a vision system show that our approach works as expected in the real world. In the future, we will consider different shapes of objects so objects may have limited reachable directions. We will also consider non-prehensile actions like pushing and dragging since some objects may need to be moved slightly to avoid collisions. Lastly, generating 3D paths of the end-effector is an interesting direction.


\bibliographystyle{IEEEtran}
\bibliography{references}

\end{document}